\documentclass[nocopyrightspace]{sigplanconf}
\usepackage{booktabs}   %% For formal tables:
\usepackage{subcaption} %% For complex figures with
\usepackage{comment,soul}
\usepackage{graphicx,tabulary}
\usepackage{amsmath,amssymb,mathrsfs,bm}
\usepackage{breakurl,authblk}
\usepackage[breaklinks]{hyperref}
\usepackage{algorithm,algpseudocode}
\usepackage{epsfig,endnotes,color,wrapfig,subcaption,xcolor}
\usepackage{subcaption}
\usepackage[font={small}]{caption}
\usepackage{setspace}
\usepackage{makecell}
\usepackage[frozencache]{minted}
\usepackage{color}
\usepackage{amsmath}
\usepackage{amssymb}
\usepackage{multirow, varwidth}
\usepackage{stfloats}
\usepackage{verbatim}
\makeatletter
\newif\if@restonecol
\makeatother
\let\oldnl\nl% Store \nl in \oldnl
\newcommand{\nonl}{\renewcommand{\nl}{\let\nl\oldnl}}
\usepackage{xr}
\usepackage[amsmath,thmmarks]{ntheorem}

\algrenewcommand\algorithmicindent{0.5em}
\algnewcommand{\IIf}[1]{\State\algorithmicif\ #1\ \algorithmicthen}
\algnewcommand{\EndIIf}{\unskip\ \algorithmicend\ \algorithmicif}

\algnewcommand{\FFor}[1]{\State\algorithmicfor\ #1\ \algorithmicdo}
\algnewcommand{\EndFFor}{\unskip\ \algorithmicend\ \algorithmicfor}

\newcolumntype{K}[1]{>{\centering\arraybackslash}m{#1}}
\newcolumntype{G}[1]{>{\centering\arraybackslash}c{#1}}

\newcolumntype{L}[1]{>{\raggedright\let\newline\\\arraybackslash\hspace{0pt}}m{#1}}
\newcolumntype{C}[1]{>{\centering\let\newline\\\arraybackslash\hspace{0pt}}m{#1}}
\newcolumntype{R}[1]{>{\raggedleft\let\newline\\\arraybackslash\hspace{0pt}}m{#1}}

\theorembodyfont{\normalfont}
\theoremseparator{.}
\newtheorem{theorem}{Theorem}%[section]

\newtheorem{definition}[theorem]{Definition}
\newtheorem{proposition}[theorem]{Proposition}

\theoremstyle{nonumberplain}

\DeclareRobustCommand\onedot{\futurelet\@let@token\@onedot}
\def\onedot{. }

\DeclareMathAlphabet{\pazocal}{OMS}{zplm}{m}{n}

\usepackage{listings}
\usepackage{setspace}
\definecolor{Code}{rgb}{0,0,0}
\definecolor{Decorators}{rgb}{0.5,0.5,0.5}
\definecolor{Numbers}{rgb}{0.5,0,0}
\definecolor{MatchingBrackets}{rgb}{0.25,0.5,0.5}
\definecolor{Keywords}{rgb}{0,0,1}
\definecolor{self}{rgb}{0,0,0}
\definecolor{Strings}{rgb}{0,0.63,0}
\definecolor{Comments}{rgb}{0,0.63,1}
\definecolor{Backquotes}{rgb}{0,0,0}
\definecolor{Classname}{rgb}{0,0,0}
\definecolor{FunctionName}{rgb}{0,0,0}
\definecolor{Operators}{rgb}{0,0,0}
\definecolor{Background}{rgb}{0.98,0.98,0.98}

\usepackage{minted}

\begin{document}

\setlength{\pdfpageheight}{\paperheight}
\setlength{\pdfpagewidth}{\paperwidth}

%% Title information
\title{Cavs: A Vertex-centric Programming Interface for Dynamic Neural Networks}         %% [Short Title] is optional;
                                        %% when present, will be used in
                                        %% header instead of Full Title.
%\titlenote{with title note}             %% \titlenote is optional;
                                        %% can be repeated if necessary;
                                        %% contents suppressed with 'anonymous'
%\subtitle{Subtitle}                     %% \subtitle is optional
%\subtitlenote{with subtitle note}       %% \subtitlenote is optional;
                                        %% can be repeated if necessary;
                                        %% contents suppressed with 'anonymous'

%% Author information
%% Contents and number of authors suppressed with 'anonymous'.
%% Each author should be introduced by \author, followed by
%% \authornote (optional), \orcid (optional), \affiliation, and
%% \email.
%% An author may have multiple affiliations and/or emails; repeat the
%% appropriate command.
%% Many elements are not rendered, but should be provided for metadata
%% extraction tools.

%% Author with single affiliation.
\author{Hao Zhang}
%\authorinfo{$^{1,2}$Hao Zhang$^{\dagger}$, $^{1,3}$Shizhen Xu$^{\dagger}$, $^{1,2}$Graham Neubig, $^{1,2}$Wei Dai, \\ $^{3}$Qirong Ho, $^2$Guangwen Yang, $^{3}$Eric P. Xing}
%{$^{1}$Carnegie Mellon University, $^{2}$Tsinghua University, $^{3}$Petuum Inc.\\ $^{\dagger}$ indicates equal contributions}
\authorinfo{Hao Zhang$^{\dagger}$, Shizhen Xu$^{\dagger}$, Graham Neubig, Wei Dai, \\ Qirong Ho, Guangwen Yang, Eric P. Xing}
           {Carnegie Mellon University, Tsinghua University, Petuum Inc.\\
           $^{\dagger}$ indicates equal contributions}
           {}

%% Paper note
%% The \thanks command may be used to create a "paper note" ---
%% similar to a title note or an author note, but not explicitly
%% associated with a particular element.  It will appear immediately
%% above the permission/copyright statement.
%\thanks{with paper note}                %% \thanks is optional
                                        %% can be repeated if necesary
                                        %% contents suppressed with 'anonymous'
\newcommand{\hao}[1]{{\color{red}{\small\bf\sf [Hao: #1]}}}
\maketitle
%% Abstract
%% Note: \begin{abstract}...\end{abstract} environment must come
%% before \maketitle command
\begin{abstract}
Recent deep learning (DL) models have moved beyond static network architectures to dynamic ones, handling data where the network structure changes every example, such as sequences of variable lengths, trees, and graphs. Existing dataflow-based programming models for DL---both static and dynamic declaration---either cannot readily express these dynamic models, or are inefficient due to repeated dataflow graph construction and processing, and difficulties in batched execution. We present Cavs, a vertex-centric programming interface and optimized system implementation for dynamic DL models. Cavs represents dynamic network structure as a static vertex function $\mathcal{F}$ and a dynamic instance-specific graph $\mathcal{G}$, and performs backpropagation by scheduling the execution of $\mathcal{F}$ following the dependencies in $\mathcal{G}$. Cavs bypasses expensive graph construction and preprocessing overhead, allows for the use of static graph optimization techniques on pre-defined operations in $\mathcal{F}$, and naturally exposes batched execution opportunities over different graphs. Experiments comparing Cavs to two state-of-the-art frameworks for dynamic NNs (TensorFlow Fold and DyNet) demonstrate the efficacy of this approach: Cavs achieves a near one order of magnitude speedup on training of various dynamic NN architectures, and ablations demonstrate the contribution of our proposed batching and memory management strategies.
\end{abstract}

%% Keywords
%% comma separated list
%\keywords{keyword1, keyword2, keyword3}  %% \keywords is optional

\section{Introduction}
\label{sec:introduction}
% deep learning,  dataflow graphs, they are effective, especially on static structures, 
Deep learning (DL), which refers to a class of neural networks (NNs) with deep architectures, is now a workhorse powering state-of-the-art results on a wide spectrum of tasks \cite{Yan:2015:HDCNN,yan2016automatic,mikolov2013efficient}. One reason for its widespread adoption is the variety and quality of software toolkits, such as Caffe~\cite{jia2014caffe}, TensorFlow~\cite{abadi2016tensorflow} and DyNet~\cite{neubig2017dynet,neubig2017fly}, which ease programming of DL models, and speed computation by harnessing modern computing hardware (e.g. GPUs), software libraries (e.g. CUDA, cuDNN~\cite{chetlur2014cudnn}), and compute clusters~\cite{zhang2015poseidon,zhang2017poseidon,cui2016geeps}. 

% reasons: they are open to a lot of graph execution optimization, and batched computation
One dominant paradigm in the training of DL models, adopted by toolkits such as Caffe and TensorFlow, uses static dataflow graphs~\cite{abadi2016tensorflow,murray2013naiad}.
These graphs represent the flow of data through computational functions, and are defined using symbolic programming~\cite{Bergstra:2011:NIPSW,abadi2016tensorflow}, once before beginning training or testing of the model.
The training of these models is performed through auto-differentiation, in which users are only required to assemble their model architectures by connecting operators using high-level language interface (e.g. Python), after which the framework will automatically derive the correct algorithm for training~\cite{bartholomew2000automatic}.
With proper optimization, the execution of these static dataflow graphs can be highly efficient.
Specifically, by separating model declaration and execution, it makes it possible for the graph to be further processed and optimized before runtime~\cite{abadi2016tensorflow}.
In addition, the evaluation of multiple data samples in a dataflow graph can be naturally batched to leverage the improved computational capability of 
%batched computation libraries (e.g. MKL, CUDA) offered by
modern hardware (e.g. GPUs), which is extremely advantageous for DL workloads~\cite{Krizhevsky:2012:NIPS}.

% dynamic NNs, dynamic declaration, their problems. 
While these static dataflow graph have major efficiency advantages, their applicability highly relies on a key assumption -- the dataflow graph (i.e. NN architecture) fixed throughout the runtime.
With the increasing complexity of the problems to be addressed, DL has been extended and applied on data with more complicated structures, such as sequences~\cite{hochreiter1997long,sutskever2014sequence}, trees~\cite{tai2015improved} and graphs~\cite{liang2016semantic}, over which the NN may conditionally choose its own computation order for specific modeling needs, i.e. the structure of its dataflow graph changes over training.
%In this case, the overhead of graph construction is constant as it only needs to be performed once before running, and the computations over multiple samples can be by nature batched (as all samples will be flow through the same graph).
%This condition breaks for \emph{dynamic models} where the dataflow graph changes over training, %in which dependency exists within one sample and the dataflow graph relies on the input data, 
%such as neural networks (NNs) computed over graph structures~\cite{tai2015improved,yan2015hd,liang2017recurrent}. %Specifically, when the model has a non-static computational workflow, such as with variably sized or even structured inputs/outputs~\cite{neubig2017dynet}, the overhead by graph construction is linear in the number of training samples, and instance-specific computational workflows prevent trivial realization of batched computation across different samples.
To better support these dynamic models, some recent frameworks~\cite{tokui2015chainer,neubig2017dynet} propose to declare a dataflow graph per sample (a.k.a. \emph{dynamic declaration}). While dynamic declaration is convenient to developers as code can basically be written in the same way as it usually is in the native programming language (e.g. Python, C++), it exhibits a few limitations.
First, programmers still have to write code to explicitly assemble the dataflow graph for each input sample, which might be nontrivial for graphs with sophisticated structures.
Second, as the graph construction needs to be performed repeatedly, its overhead grows linearly with the number of training instances, preventing the application of complex static graph optimization techniques (in fact, graph construction takes longer time than the computation in some frameworks~\cite{looks2017deep}, see \S\ref{sec:graph_construction}).
Finally, since each sample owns a dataflow graph specifying its unique computational pattern, batching together similarly shaped computations across instances is non-trivial.
Without batching operations, the computation is inefficient due to its lack of ability to exploit modern computational hardware, and while some progress has been made in recent research \cite{neubig2017fly,looks2017deep}, how to automatically batch the computational operations from different graphs remains a difficult problem.

% Present Cavs
% 1. motivation
% 2. Cavs' design
To address these challenges, we present Cavs, a new programming interface for dynamic NNs, and a system implementation with optimization strategies tailored to it. Cavs leverages the recurrent and recursive nature of dynamic NNs. Instead of declaring a dataflow graph per sample, it alternatively decomposes a dynamic dataflow graph as two components: one static vertex function $\mathcal{F}$ that is only declared (by the user) and optimized once, and an input graph $\mathcal{G}$ that is instance-specific and not used until runtime. Thereby, the workflow of training a dynamic NN can be represented as scheduling the execution of $\mathcal{F}$ following the structure of the input graph $\mathcal{G}$. Cavs combines the best of symbolic construction of dataflow graphs for DL~\cite{abadi2016tensorflow,Bergstra:2011:NIPSW} with the vertex-centric model~\cite{gonzalez2012powergraph} in graph computing: it only requires users to define $\mathcal{F}$ symbolically by ``thinking locally like a vertex''~\cite{tian2013think}. Cavs will perform auto-differentiation, schedule the function execution following the dependency reflected by $\mathcal{G}$, and guarantee efficiency and correctness. It also inherits the flexibility of symbolic programming, i.e. users are allowed to declare multiple vertex functions to express more dynamics, or connect static dataflow graphs with dynamic ones to construct more complex NN architectures.

% 3. advantages
Cavs demonstrates a few advantages over other programming models. It simplifies user programs and avoids the overhead of repeated dataflow graph construction. Moreover, this vertex-centric model naturally exposes opportunities for batched computation: we introduce a simple batching policy in Cavs' scheduler to parallelize the execution of $\mathcal{F}$ on multiple vertices during the evaluation of a batch of samples with different input graphs (\S\ref{sec:scheduling}), and a novel memory management mechanism to guarantee the memory coalescing (\S\ref{sec:memory_management}). Together they yield significant performance improvements. Compared to dynamic declaration, as the dataflow graph encoded by the vertex function is static throughout the runtime, it can benefit from various static graph optimizations~\cite{abadi2016tensorflow,chen2015mxnet,caffe2, xla}, such as lazy batching, streaming, and kernel fusion (\S\ref{sec:graph_execution}), which would otherwise be less effective on the scenario of dynamic declaration because of the repeated preprocessing/optimization cost (see \S\ref{sec:related_work}).

%Existing optimizations such as kernel fusion from static frameworks, and memory coalescing for efficient batch evaluation, can also be easily incorporated. %Cavs purposes Gather/Scatter primitives for data communication on the dependency edge. To make it expressive to stack more layers, Push/Pull primitives are also introduced for the inter-layer data communication.
%The vertex-centric model naturally exposed the parallelism within one batch and within one sample and therefore the batching can be easily applied. In the same time, Cavs narrows the graph optimization into the optimization of vertex dataflow graph. Existing optimizations such as kernel fusion in static frameworks can be easily applied. Cavs purposes Gather/Scatter primitives for data communication on the dependency edge. To make it expressive to stack more layers, Push/Pull primitives are also introduced for the inter-layer data communication.

% 4. implementation and empirical performance.
We implement Cavs as an additional layer pluggable to most existing DL frameworks to enhance their support for dynamic NNs. To evaluate its performance, we compare Cavs to TensorFlow Fold~\cite{looks2017deep} and DyNet~\cite{neubig2017dynet,neubig2017fly}, two state-of-the-art systems supporting dynamic NNs and dynamic batching. We focus our experiments on GPU training, and verify that both Fold and DyNet suffer from substantial overhead caused by repeated graph preprocessing or construction, which is bypassed by Cavs (\S\ref{sec:graph_construction}). In terms of overall performance, on static NNs, Cavs demonstrates equivalent or slightly better performance than Fold and DyNet, while on several dynamic NNs with notably difficult-to-batch workloads (e.g. Tree-LSTM~\cite{tai2015improved} and Tree-FC~\cite{looks2017deep}), Cavs demonstrates near one  order of magnitude speedups across various dataset and hyper-parameter settings (\S\ref{sec:overall_performance}). We further investigate the key contributing factors to the performance: Cavs benefits from not only a better memory management strategy, but also graph execution optimizations which are originally designed for static dataflow graphs and perhaps less useful in dynamic declaration.

\begin{figure*}[bpt]
\centering
\begin{subfigure}{0.32\textwidth}
  \centering
  \begin{minted}[%linenos,
               numbersep=0pt,
               %gobble=25,
               frame=single,
               framesep=-0.5mm,
               fontfamily=helvetica,
               escapeinside=||,
               fontsize=\footnotesize,
               mathescape=true]{cpp}
  /* (a) static declaration */
  // all samples must share one graph
  declare a |static| dataflow graph |$\mathcal{D}$|.
  for |$p = 1 \rightarrow P$|:
    read the |$p$|th data batch |$\{x_k^p\}_{k=1}^K$|.
    batched |computation|: |$\mathcal{D}(\{x_k^p\}_{k=1}^K)$|.
\end{minted}
\end{subfigure}
\begin{subfigure}{0.32\textwidth}
  \centering
  \begin{minted}[%linenos,
               numbersep=0pt,
               %gobble=25,
               frame=single,
               framesep=-0.5mm,
               fontfamily=helvetica,
               escapeinside=||,
               fontsize=\footnotesize,
               mathescape=true]{cpp}
  /* (b) dynamic declaration */
  for |$p = 1 \rightarrow P$|:
    read the |$p$|th data batch |$\{x_k^p\}_{k=1}^K$|.
    for |$k = 1 \rightarrow K$|:
        declare a dataflow graph |$\mathcal{D}_k^p$| |for| |$x_k^p$|.
        single-instance |computation|: |$\mathcal{D}_k^p(x_k^p)$|.
\end{minted}
\end{subfigure}
\begin{subfigure}{0.35\textwidth}
  \centering
  \begin{minted}[%linenos,
               numbersep=0pt,
               %gobble=25,
               frame=single,
               framesep=-0.5mm,
               fontfamily=helvetica,
               escapeinside=||,
               fontsize=\footnotesize,
               mathescape=true]{cpp}
  /* (c) our proposed vertex-centric model */
  declare a symbolic vertex function |$\mathcal{F}$|. 
  for |$p = 1 \rightarrow P$|:
    read the |$p$|th data batch |$\{x_k^p\}_{k=1}^K$|.
    read their associated graphs |$\{\mathcal{G}_k^p\}_{k=1}^K$|.
    compute |$\mathcal{F}$| over |$\{\mathcal{G}_k^p\}_{k=1}^K$| with inputs |$\{x_k^p\}_{k=1}^K$|.
\end{minted}
\end{subfigure}
\vspace{-8pt}
\caption{\small The workflows of (a) static declaration, (b) dynamic declaration, (c) Cavs' vertex-centric programming model.}
\label{fig:programming_models}
\end{figure*}

\begin{figure*}[bpt]
%\centering
%\raggedright
\vspace{-15pt}
\begin{subfigure}{0.33\textwidth}
%\raggedleft
%\hspace{5pt} 
\includegraphics[width=\textwidth]{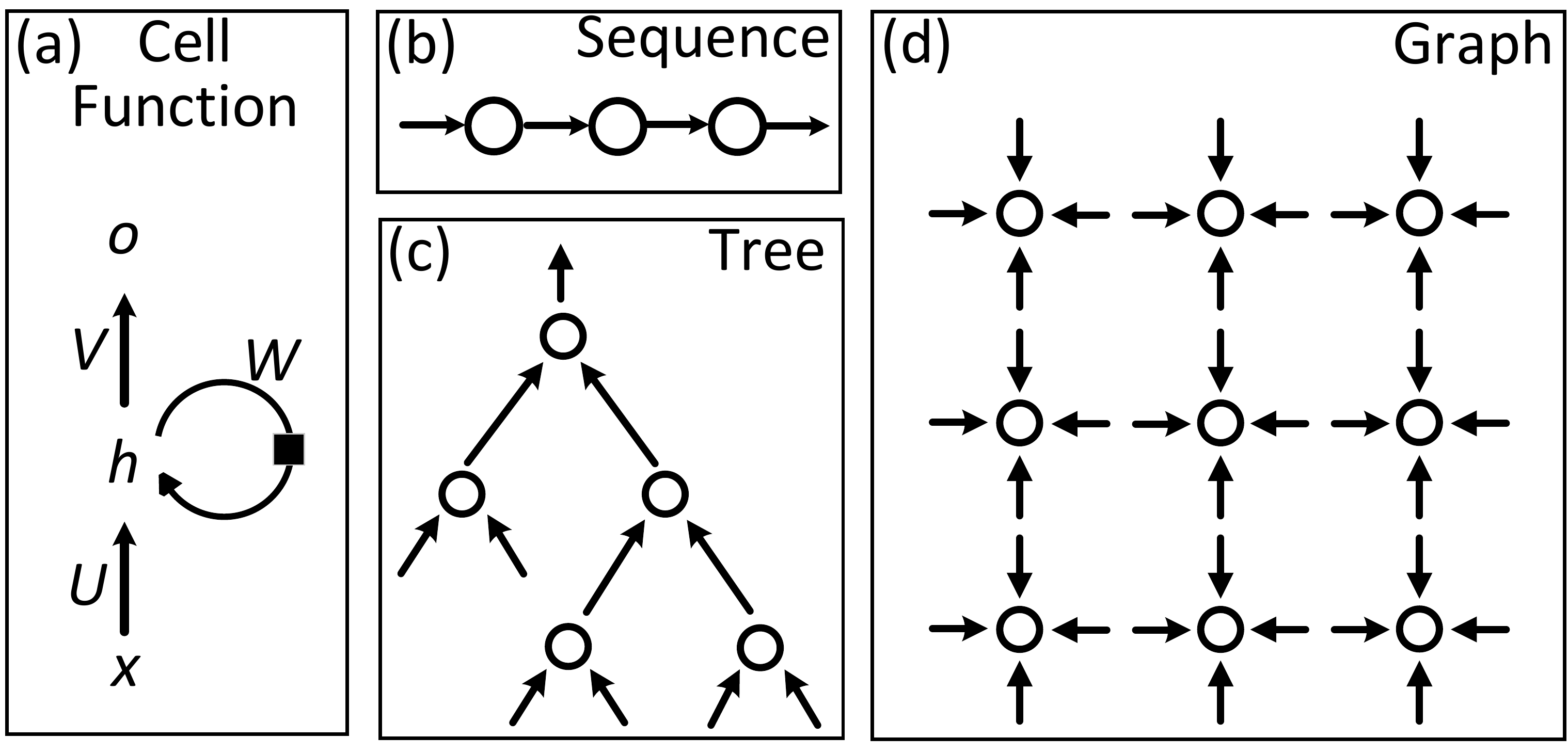}
\end{subfigure}  
\begin{subfigure}{0.46\textwidth}
\scriptsize
%\raggedleft
%\hspace{7pt} 
\begin{tabular}{ |K{2.0cm}|K{2.0cm}|K{1.45cm}|K{0.85cm}|K{1.35cm}|K{1.4cm}|} 
\hline
%\multirow{2}{*}{\textbf{Model}} & \multirow{2}{*}{\textbf{Frameworks}} & \multicolumn{2}{c|}{\textbf{Expressiveness}}  & \multicolumn{2}{c|}{\textbf{Batching}} & \multirowcell{2}{\textbf{Graph Cons.} \\\textbf{Overhead}} & \multirowcell{2}{\textbf{Graph Exec.} \\\textbf{Optimization}} \\
%\cline{3-6}
%& & static & dynamic & static& dynamic & & \\
\textbf{Model} & \textbf{Frameworks} & \textbf{Expressiveness}  & \textbf{Batching} & \textbf{Graph Cons. Overhead} & \textbf{Graph Exec. Optimization} \\
\hline
\hline
static declaration & Caffe, Theano, TensorFlow, MxNet & $\times$ &  $\times$  & low & beneficial\\
\hline
dynamic declaration (instant evaluation) & PyTorch, Chainer & $\surd$ & $\times$  &  N/A & unavailable \\
\hline
dynamic declaration (lazy evaluation) & DyNet & $\surd$ & $\surd$ &  high & not beneficial\\
\hline
Fold & TensorFlow-Fold &  $\surd$ & $\surd$ &  high & unknown\\
\hline
Vertex-centric & Cavs &$\surd$ & $\surd$ & low & beneficial\\
\hline
\end{tabular}
\end{subfigure}
\vspace{-5pt}
\caption{\small Left (a)-(d): A cell function shown in (a) could be applied on different structures such as a (b) chain (c) tree, or (d) graph. Right table: the landscape of existing programming models for dynamic NNs, and their advantages and disadvantages (see \S\ref{sec:programming_dynamic} and \S\ref{sec:related_work}).}
\label{fig:recurrent}
\vspace{-15pt}
\end{figure*}

\vspace{-20pt}
\section{Background}
\label{sec:background}
%In this section, we brief the background of DL and dataflow graph representations. We then introduce dynamic NNs, summarize existing programming models for such problems~\cite{looks2017deep,neubig2017dynet,pytorch,chainer}, and discuss their limitations.
%\subsection{Deep Learning as Dataflow Graphs}
%\label{sec:dataflow_graph}
DL is distinguished from other ML algorithms mainly by its use of deep neural networks, a family of ML models with many interconnected layers, each composed of various mathematical operations (e.g. $+, -, \texttt{sigmoid}, \texttt{matmul}$). Before a DL model can give predictions, it is usually trained by stochastic gradient descent, an iterative process in which gradients are calculated through backpropagation~\cite{rumelhart1988learning}.
%\shizhen{[DEL]At each iteration of the process, a chunk of input data, usually represented as a multi-dimensional array (tensor), is forwarded through the network following its layered structure, and then backward in reverse to compute the parameter updates, repeatedly until the model parameters have been updated to some state that is regarded as optimal.} 
There is a natural connection between directed graphs and NNs: we can map the graph nodes to the computational operations or parameters in NNs, and let the edges indicate the direction of the data being passed between the nodes. In this case, we can represent the process of training NNs as batches of data flowing through computational graphs, i.e. \emph{dataflow graphs}~\cite{Bergstra:2011:NIPSW, abadi2016tensorflow, neubig2017dynet}. 
% As stated in the introduction, this view of NNs has many advantages and have become the major form to represent DL models. 
\begin{comment}
%in which the users first declares their NNs by assembling a dataflow graph, and then invokes the training/inference by feeding data samples to the graph. Dataflow graph representation presents a few advantages: It only requires DL programmers to express the forward computational graph of the neural network by connecting a few primitive operators, the backward gradient derivation can be handled by auto-differentiation; It helps separate the model declaration from execution, therefore it allows users to write \emph{symbolic programs}~\cite{abadi2016tensorflow,Bergstra:2011:NIPSW} using any high-level language (e.g. python); for framework developers, as long as the program can be complied as a valid dataflow graph representation, they can later apply various graph-based optimizations over it for improved efficiency. 
\end{comment}
% explain why batching is essential to the performance of
% graph construction, batched operation are essential to the performance of DL programs

Figure~\ref{fig:programming_models}(a) summarizes the programming model derived from these dataflow graphs, which is named as \emph{static declaration} and has been adopted in many DL frameworks~\cite{Bergstra:2011:NIPSW,abadi2016tensorflow,chen2015mxnet}. Without ambiguity, we use $\mathcal{D}$ to denote both the dataflow graph itself and the computational function implied by $\mathcal{D}$. 
On one hand, we note its execution is highly efficient as the computation over multiple samples is batched -- at each iteration $p$, a batched tensor of $K$ samples $\{x_k^p\}_{k=1}^K$  is fed to $\mathcal{D}$, and the computation is executed in a single pass, allowing for efficient use of memory caches or parallelized computation. 
%\shizhen{parallelized by leveraging batched computational libraries/hardware}. 
On the other hand, this paradigm relies on a key assumption: the dataflow graph $\mathcal{D}$ is static for all samples and fixed throughout the computation. Hence, $\mathcal{D}$ will only be declared once with a constant graph construction/optimization overhead; all samples share the same computational pattern specified in $\mathcal{D}$, so the computation of different samples can be \emph{by nature} batched by simply expanding the input with a batch dimension.
% However, static graph assumption breaks
Though static declaration is effective on a wide range of NN models, such as convolutional neural networks (CNNs) over fixed-size images, it is much more difficult to apply to graphs with dynamically changing structures, some examples of which are shown in the next section.

%\vspace{-8pt}
\subsection{Dynamically Structured Computational Graphs}
%\gn{I'm not sure whether ``recurrent'' or ``recursive'' is better here. I feel ``recurrent'' is strongly associated with sequences.}
%With the increasing complexity of the problems to be addressed, 
Modern DL has been developed and applied extensively over data with more complicated structures, e.g. data structured as sequences, trees and graphs,
which are required to tackle practical problems such as machine translation~\cite{sutskever2014sequence,tai2015improved}, questionanswering~\cite{tan2015lstm}, and semantic image segmentation~\cite{yan2016combining,liang2016semantic}.
As a concrete example of dynamic NNs, we will use recurrent and recursive neural networks (RNNs)~\cite{elman1990finding,hochreiter1997long,socher2013recursive}. RNNs are a class of NNs generally applied on modeling structured inputs or outputs, e.g. sequences or graphs. At the core of an RNN is a cell function with trainable parameters. It will be dynamically applied at different places of the input structure, and optionally produce an output if needed. Figure~\ref{fig:recurrent}(a) illustrates such a cell function: it takes an input element $x$, forwards it through a few mathematical transformations, and generates some intermediate state $h$ and an output $o$. Depending on what transformations are applied, different variants of RNNs have been derived, such as long-short term memory units (LSTM)~\cite{hochreiter1997long} and gated recurrent units (GRU)~\cite{chung2014empirical}. However, the internals of the cells themselves are secondary; the dynamics of the net as a whole are mainly reflected by the structures that the NN works on.
%\hao{figure2(a): remove the circles, change $s$ to $h$, change ``'chain' to ``sequence''...}

\noindent \textbf{Sequence RNNs.}
When the input to the RNN are sequences (e.g. sentences) as in Figure~\ref{fig:recurrent}b, the cell function is applied across all elements of the sequence. At each step $t$, it takes the element $x_t$ (e.g. a word) from the input sequence, and the state variable $h_{t-1}$ maintained by the model at step $t-1$. It computes an output $o_{t}$, and a new state $h_t$ that will be used by the next step $t+1$. Hence, This sequence RNN encodes not only the data values, but also the dependencies present in the sequence. If represented as a dataflow graph, the graph exhibits a chain structure. As the input or output sequences usually have variable length (e.g. translating an arbitrary-length English sentence into Chinese), the dataflow graph needs to be dynamically changed, i.e. the steps of the chain must be adapted to fit the length of the input or output.

%As the input sequence $x$ is usually variable-length, or even with internal structures, the DL models usually need to be adapted accordingly to encode such structured dependencies, for example, compute over some sophisticated graphs. A number of DL models have been developed with this desired properties, which we briefly discuss below in the context of sequence prediction problems.

%Figure.\ref{} presents the computational graph of a sequential LSTM that has been broadly applied for machine translation. At each time step $t$, its cell function reads a word from the input sentence, update its internal states $s_t$ to $s_{t+1}$, and outputs a translated word $y_t$. The parameters of the cell function is shared across all steps $t = 1. \dots, T$, and updated by minimizing a desired loss w.r.t. the parameters and backpropagating the gradients through all time steps, a.k.a. backpropagation through time (BPTT). 

\noindent \textbf{Tree-structured RNNs.}
Further, RNNs can be enhanced to model data with more complex structures suited for downstream tasks. For example, tree-structured RNNs (Tree-RNNs, Figure~\ref{fig:recurrent}c), have been used to classify the sentiment of sentences~\cite{pang2002thumbs} given an associated binary tree representing the sentence structure~\cite{tai2015improved,socher2011parsing}. In this case, a leaf of the tree maps to a word of the sentence, an internal node corresponds to a multi-word phrase. To process this structure, the cell function scans the tree recursively, starting from leaf nodes until reaching the root. At the node $t$, it computes the state $h_t = f(h_{t_l}, h_{t_r}, x_t)$, where $x_t$ is the input to the node, and $h_{t_l}, h_{t_r}$ are the states of its left and right children, respectively. As the tree structure vary from example to example, the dataflow graph of a Tree-RNN is highly dynamic. 

\noindent \textbf{Graph-structured RNNs.}
Similarly, RNNs can be extended to compute over more general graphs, such as N-ary trees or graphs (Figure~\ref{fig:recurrent}d), as long as their parameters are learnable. In fact, various NNs have been developed toward having more dynamic workflows~\cite{liang2016semantic,tai2015improved}, and proven quite effective because of their ability to encode structured information. 
While we take RNNs as examples for explanation, we note there are many other dynamic NNs in the literature or production with their dynamics reflected in various perspectives: variably sized inputs/outputs~\cite{sutskever2014sequence,bahdanau2014neural,elman1990finding,hochreiter1997long, dyer2015transition, buckman2016transition}, variably structured inputs/outputs~\cite{socher2011parsing,tai2015improved,liang2016semantic}, or with nontrivial inference algorithms~\cite{graves2006connectionist,zheng2015conditional,gormley2015approximation,kong2015segmental}. 
%We next discuss existing programming models and training strategies for them.

\vspace{-8pt}
\subsection{Programming Dynamic Dataflow Graphs}
%\vspace{-5pt}
\label{sec:programming_dynamic}
%\gn{I haven't checked this whole section carefully. Will do so pending discussion w/ Hao.}
%In order to model the structure information associated with data values, the NN may conditionally choose its own computation order according to its input or output. 
As the assumption in \S\ref{sec:background} no longer holds for dynamic structures, static dataflow graphs in their original form cannot be used.
There are currently two remedies to this problem: expanding the graph programming language to allow it to explicitly include controls structure necessary to implement these applications, or forgo the efficiency gains afforded by static dataflow graphs and instead use a dynamic declaration framework that reconstructs the graph for every training example.
We explain below and summarize them in Figure~\ref{fig:recurrent}.

\noindent \textbf{Static declaration.}
\emph{Static unrolling}~\cite{abadi2016tensorflow} is a standard way to express sequence RNNs with fixed steps.
% --- it simply declares a single fixed-structured dataflow graph by unrolling the cell function a fixed number of steps. 
To handle variable-length data, it declares an RNN that has number of steps equal with the length of the longest sequence in the dataset. It then appends zeros at the end of other sequences to have equal length, and feeds them in batches to the dataflow graph for computation. Static unrolling enables batched computation of multiple sequences, but obviously results in substantial unnecessary computation.%
\footnote{It is also possible to split sentences into several buckets of different lengths, which alleviates this problem somewhat but adds some degree of code complexity and is not a fundamental solution.}
% \emph{Bucketing} improves over static unrolling -- instead of padding all inputs to have the same size, it buckets them according to their length (sequences not bucketable are zero-padded to be put into the nearest bucket). It then creates a fixed-step RNN (i.e. a dataflow graph) for each bucket, and performs batched computation therein. Compared to static unrolling, bucketing saves considerable computation, but introduces a bucketing process that adds extra complexity. 
%Both static unroll and bucketing are efficient because they batched computation. 
\emph{Dynamic unrolling} implements basic control flow functionality within static graphs, allowing for the declaration of graph operators similar to \texttt{while} loops. At each iteration of the training, the cell function of the RNN will be executed a conditional number of times determined at runtime by the length of the longest sentence in the batch. It then pads the sequences in the batch and perform batched computation, it waste computational resources.
Notably, both of these methods essentially cannot support more complex structures than sequences.

\noindent \textbf{Dynamic declaration.}
%\gn{Is TensorFlow fold a dynamic declaration framework? It's hard to say. I think we should discuss this.}
%\shizhen{To express those complicated dynamic dataflow graphs, existing frameworks~\cite{abadi2016tensorflow,neubig2017dynet} resort to the dynamic declaration model.} 
%To overcome the inflexibility of static declaration, \emph{dynamic declaration}~\cite{neubig2017dynet}, illustrated in Figure~\ref{fig:programming_models}b, is a recently developed programming model suitable for dynamic NNs.
%While existing frameworks are all designed originally for static computational flows, there have been a surge of interest in adapting them to better fit dynamic workflows, such as PyTorch~\cite{}, which dynamically declares static graphs for each data sample, as well as DyNet~\cite{} and TensorFlow-Fold~\cite{}, that further explores the potential opportunities to batch operations from multiple static graphs.
Dynamic declaration is able to express dynamically varying dataflow graphs, by creating a unique dataflow graph for each sample according to its associated structure. %at the cost of losing many opportunities for batched computation of multiple samples.
%As a programming model, dynamic declaration is compatible with most DL frameworks, as it demands little change on low-level implementations of the original framework. 
It however requires users to explicitly write (more) code to build a dataflow graph for each input graph, which is nontrivial for graphs with sophisticated structures. 
As the dataflow graphs are always changing, it can hardly benefit from well-established dataflow graph optimization techniques (\S\ref{sec:graph_execution}) -- we will have to perform graph processing/optimization for each dataflow only for a single sample, but incorporating the optimization itself has an overhead.
%s carried out once for static dataflow graph, has to be carried out repeatedly for each mini-batch and therefore is not as beneficial. 
More importantly, as we are unable to naturally batch the computation of different sample, single-instance training would be very inefficient in the absence of batched computation. 
At the backend, since a dataflow graph needs to be constructed per sample, the overhead is linearly increasing with the number of samples, and sometimes yields downgraded performance~\cite{looks2017deep} (\S\ref{sec:graph_construction}), even for frameworks with optimized graph construction implementations~\cite{neubig2017dynet}. 
%We compare the five major methods for dynamic dataflow graphs in Table~\ref{}.
%RNN is suitable for sequence modeling as it not only encodes each element of the sequence, but also the long-term dependencies specified by the structure of the data.
%Training RNNs is usually computationally intensive, as the cell function usually involves dense matrix computation, and needs to be repeated as many times as the length of the input for each data point. Therefore, accelerating the computation by batching elementwise operations as much as possible is essential for efficient training.
%\subsection{Challenges with Dynamic Data Flow Graphs}
%\noindent \textbf{Efficiency}
%\noindent \textbf{Expressiveness}.
%While dynamic declaration and imperative programming model can meet the demand of dynamically varying dataflow graph, the efficiency remains the major issue. This is because: 1) batched execution, which is a natural feature of static dataflow graph, is not that evident and 2) dataflow graph optimization, which is carried out once for static dataflow graph, has to be carried out repeatedly for each mini-batch and therefore is not as beneficial. 

Tensorflow Fold~\cite{looks2017deep} and DyNet~\cite{neubig2017fly} go one step further and perform dynamic batching for dynamic dataflow graphs. Fold turns dynamic dataflow graphs into a static control flow graph to enable batched execution, but introduces a complicated functional programming-like languages and a large graph preprocessing overhead. DyNet proposes an auto-batching strategy that searches for batching opportunities by profiling every fine-grained operator, while this step itself has non-negligible overhead, and loses the opportunities of graph-level optimizations.
%In summary, we present a landscape and comparison of existing programming models for dynamic NNs in the table in Figure~\ref{fig:recurrent}. 
There are also some ``imperative'' frameworks, such as PyTorch~\cite{pytorch} and Chainer~\cite{chainer} that allow users to construct dynamic NNs but performs instant evaluation of each user expression. As model construction and execution are coupled, they are usually difficult to perform dynamic batching or graph optimization. Overall, they are still far from efficient when handling dynamic NNs.
We next describe our proposed vertex-centric programming model to overcome the aforementioned limitations. 
%Dynamic dataflow graphs means each sample needs a new data-flow graph. Static declaration, which is based on the assumption that each sample shares the same data-flow graph is invalid for such a scenario. 

\vspace{-10pt}
\section{Cavs Design and Optimization}
%\vspace{-5pt}
\label{sec:design}
Our motivation comes from several key principles ML developers usually comply with to ensure the feasibility and learnability of the model during their design of dynamic NNs.
We note most dynamic NNs are designed to exhibit a recursive structure (e.g. sequence RNN, Tree-RNN), or a combination of static and recursive structures (e.g. LRCN~\cite{donahue2015long,andreas2016neural}, attention~\cite{xu2015show}), or even a combination of different recursive structures (e.g. encoder-decoder RNNs~\cite{sutskever2014sequence}). Within one such structure, a function is dynamically applied over instance-specific graphs, and every vertex of the graph usually interacts in a same way with it neighboring vertices following the function. 
The computational function itself, however, is usually static and parameterized by fixed learnable parameters.  
%If we take the Tree-LSTM~\cite{tai2015improved} as an example, we note it always has a single parameterized cell function that will be evaluated at each node of a tree to ensure the model parameters are trainable, instead of having a specialized cell function with an independent set of parameters for each node of the tree.
%Moreover, when the cell function scans at d11ifferent vertices of the graph, it is designed to always apply the same transformations over the child nodes of the current vertex, regardless of which node of the graph it is working on.
%\gn{This might be too strong. Socher et al.'s recursive neural networks use a different parameter matrix for each transformation, which falls between having a single operation and having a different option for each node.}

%There might exist other possible design choices, for example, for each node in a Tree-LSTM, customizing the way it interacts with its child nodes, yet this usually results in the model being unable to capture the patterns we intend to learn (not learnable), or inapplicable on various (and potentially arbitrary) input structures (because we then have to customize for each incoming tree, which is apparently impossible).

\begin{figure}[bpt]
\centering
\includegraphics[width=0.45\textwidth]{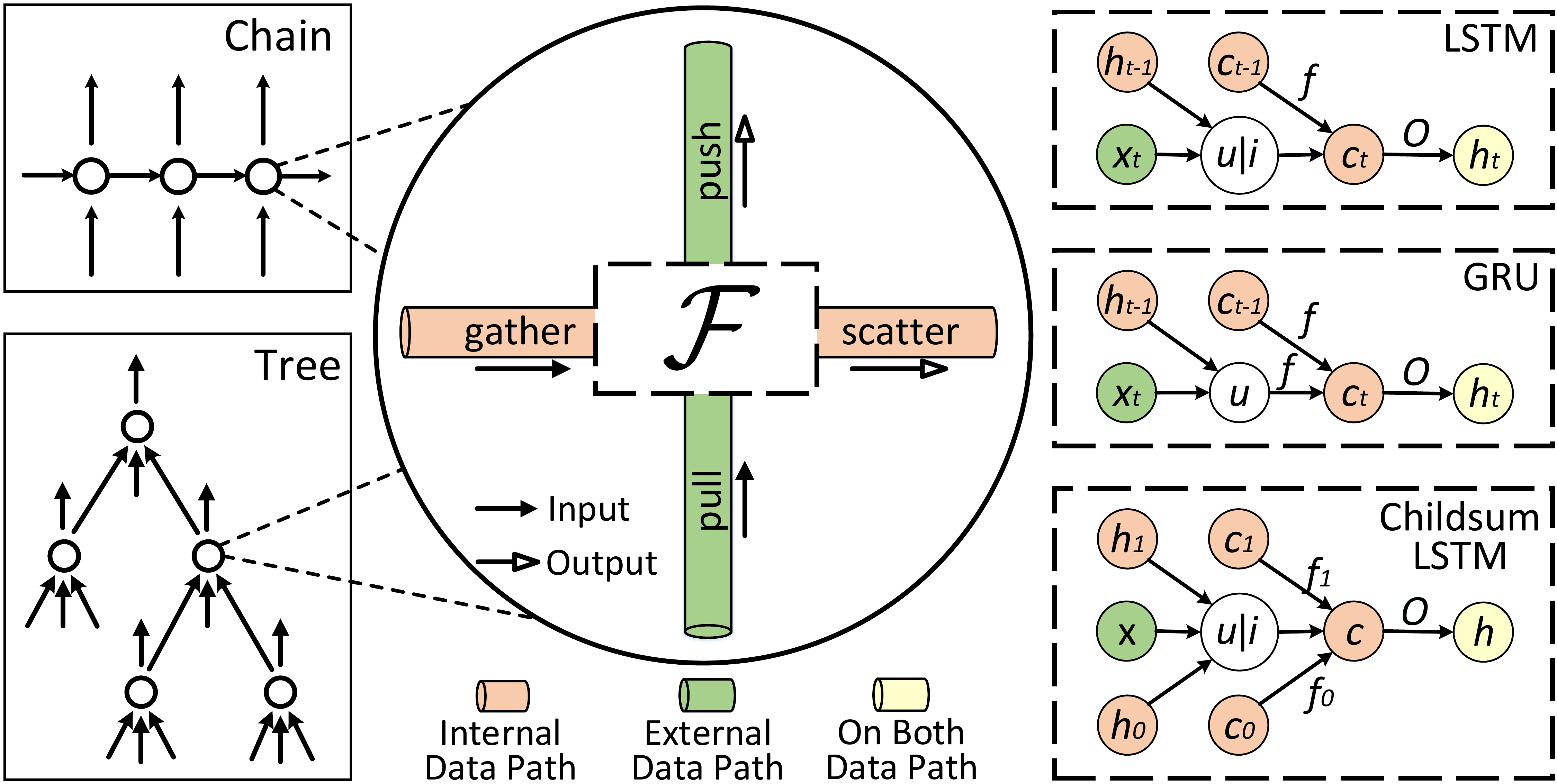}
\vspace{-5pt}
\caption{\small Cavs represents a dynamic structure as a dynamic input graph $\mathcal{G}$ (left) and a static vertex function $\mathcal{F}$ (right).}
\label{fig:cavs_concept}
\vspace{-10pt}
\end{figure}

%\gn{Reading this made me think of Neural Module networks: \cite{andreas2016neural}. This might be worth a cite.}
This observation motivates us to design a new programming model, called Cavs, that combines the best of dataflow graphs with the vertex-centric model in graph computing. As a comparison, we present the workflow of Cavs in Figure~\ref{fig:programming_models}c. For clarity, we will use the following terminology and notation in the rest of the paper: we call the instance-specific structure associated with the input sample as an \emph{input graph}, and notate it as $\mathcal{G}$, and a node in that graph as a \emph{vertex}, to be distinguished from a dataflow graph $\mathcal{D}$ and the nodes (which are usually operators or variables) therein. 
Figure~\ref{fig:cavs_concept} illustrates the concept of this vertex-centric programming model. To describe an aforementioned dynamic structure, different from dynamic declaration, which requires users to manually declare dataflow graphs for each sample according to its associated graph, Cavs instead directly takes it as an input argument.
%\gn{``instead directly takes it as an input argument'' is a little bit confusing maybe. My interpretation is at the 2 of training, we define a set of vertex functions that can be applied to any node in the graph, and take in inputs and produce outputs, but do not describe the data flow graph explicitly. The user then passes in input graph for every training example, which can \emph{only} use these pre-defined arguments. If this understanding is correct, I think this part should be reworded a bit.}
To be aware of what computation shall be performed, Cavs requires users to implement a simple vertex function $\mathcal{F}$ by ``thinking like a vertex'', informing the framework how one vertex in a dynamic NN will interact with its connected vertices (if these is any). In $\mathcal{F}$, users can utilize conventional DL operators to assemble a symbolic construct that will be evaluated dynamically following the structure of $\mathcal{G}$, while Cavs will ensure the correctness and efficiency. Therefore, a vertex function $\mathcal{F}$, together with an input graph $\mathcal{G}$, implicitly encodes a recurrent dataflow graph, which maps to a subgraph of the implicit full dataflow graph of the model that may needs to be explicitly declared in traditional programming models.
For convenience of notations, we will call any part of the structure that cannot be encoded by $\mathcal{F}$ and $\mathcal{G}$ as \emph{external to} $(\mathcal{F}, \mathcal{G})$, and vice versa. Cavs allows users to connect any external static dataflow graph to a dynamic structure encoded by $(\mathcal{F}, \mathcal{G})$ to express various model architectures (e.g. connecting a CNN to an RNN), or declare multiple vertex functions for different structures, and connect them appropriately to express more complex models (e.g. an encoder-decoder LSTM network).

While it is still necessary to create an I/O function to read input graphs for each sample, this must be done in any models, and only once before training commences, which means that it can be shared across epochs or even training runs. Cavs no longer requires users to construct the full dataflow graphs for each sample by themselves.
%therefore it significantly simplifies users' program.
%\gn{I'm not sure if this is true. Why would this be any easier?}
As repeated graph construction is bypassed, the overhead will also be avoided.
%\gn{How about this ``And while it is still necessary to create an input graph for every training instance, this must only be done once before training commences, which means that it can be shared across epochs or even training runs.''}
With this vertex-centric model, Cavs transforms the problem of evaluating multiple dataflow graphs with different structures~\cite{looks2017deep,neubig2017fly} into a simpler form -- scheduling the execution of the vertex functions following the input graphs. 
%The formal problem, which is usually called \emph{dynamic batching}, has been explored~\cite{looks2017deep,neubig2017fly} and prove intrinsically difficult. 
For the later problem, we can easily batch the execution of $\mathcal{F}$ over multiple vertices at runtime (\S\ref{sec:scheduling}), leveraging the batching computational capability of modern hardware. Moreover, as the vertex function itself maps to a static symbolic dataflow graph, it is open and can benefit from various graph optimization techniques originally developed for static declaration, such as kernel fusion, streaming, and our proposed lazy batching, which might not be effective in the scenario of dynamic declaration. 
%
%\gn{Perhaps the next two sentences are not necessary?}
%Overall, Cavs can be seen as a vertex-centric and much simpler representation that helps encode the recurrent and instance-specific structures of an NN. It is compatible with any DL framework based on symbolic declaration.
We next describe Cavs' APIs.

%\vspace{-4pt}
\subsection{Programming Interface}
%\vspace{-5pt}
\label{sec:api}
Besides the generic math operators used to declare the computation, Cavs exposes four symbolic APIs for users to specify how the messages shall be passed between vertices in their vertex functions: \texttt{gather}, \texttt{scatter}, \texttt{pull}, \texttt{push}.
\begin{itemize}
\vspace{-5pt}
\item \texttt{gather(child\_idx)}: \texttt{gather} accepts an index of the child vertices, gets the child content from gather/scatter buffer and returns a list of symbols that represent the output of these vertices.
%When \texttt{child\_idx} is null, \texttt{gather} behaves like \texttt{gather(all)}.
\vspace{-5pt}
\item \texttt{scatter(op)}: \texttt{scatter} is a reverse API of \texttt{gather}, and has a symbol \texttt{op} as its input argument. Scatter will set the output of current vertex to gather/scatter buffer. 
%as \texttt{op} -- if any other node tries to \texttt{gather} the output of this vertex, \texttt{op} will be returned.
\vspace{-5pt}
\end{itemize}
\texttt{gather} and \texttt{scatter} resemble the GAS model in graph computing~\cite{gonzalez2012powergraph} -- both are vertex-centric APIs that help users express the overall computational patterns by thinking locally like a vertex: \texttt{gather} receives messages from dependent vertices, while \texttt{scatter} updates information to parent vertices.
But note several key differences: (1) \texttt{gather} and \texttt{scatter} here are fully symbolic -- \texttt{gather} allows backpropagation through it; (2) In graph computing, all nodes interact with their connected nodes in the same way following a user-specified \texttt{apply} function, while in dynamic NNs, a vertex usually interacts differently with its different child vertices, specified by the symbolic programs (between the call of \texttt{gather} and \texttt{scatter}) in the vertex function; (3) In graph computing, a vertex of a graph always interacts with other vertices of this graph, while in DL, the vertex of a dynamic NN usually takes input from not only the internal of the structure expressed by $\mathcal{F}$ and $\mathcal{G}$ (internal data path in Figure~\ref{fig:cavs_concept}), but also from the external of it, e.g. a step in an RNN can take inputs from a CNN feature extractor or some external I/O (external data path in Figure~\ref{fig:cavs_concept}). In this case, \texttt{gather} and \texttt{scatter} are insufficient to express such semantics. Cavs therefore provides another two APIs:
\begin{itemize}
\vspace{-5pt}
\item \texttt{pull()}: \texttt{pull} grabs inputs from the external of the current dynamic structure, e.g. another NN, or some I/O. 
%It receives an argument \texttt{input\_idx} that specifies a list of indices of its external input, and returns a list of corresponding symbols. When \texttt{input\_idx} is null, it behaves like $\texttt{pull(all)}$.
\vspace{-5pt}
\item \texttt{push(op)}: \texttt{push} is thus the reverse of \texttt{pull} that sets the output of the current vertex as \texttt{op}. If this vertex is \texttt{pull}ed by others, the content of $\texttt{op}$ will be returned.
\vspace{-5pt}
\end{itemize}
With appropriate indexing, $\texttt{push}$ and $\texttt{pull}$ connect a vertex inside a dynamic structure expressed by $(\mathcal{F}, \mathcal{G})$ to other connectors external to $(\mathcal{F}, \mathcal{G})$, such as another dynamic structure, or another static dataflow graph.
With these four APIs, we present in Figure~\ref{fig:treeLSTM} an example user program how the $N$-ary child-sum Tree-LSTM~\cite{tai2015improved} can be simply expressed by using them and other mathematical operators. %\hao{add a sentence: all pull can be connected to external ops, with index.}
%\gn{Clarification again about decoding, for example, in seq2seq models. How would you decide when to stop generating output?}

\begin{comment}
\begin{figure}
\caption{$\mathcal{F}$}
\begin{minted}[%linenos,
               numbersep=0pt,
               %gobble=25,
               frame=single,
               framesep=-2mm,
               fontfamily=tt,
               escapeinside=||,
               fontsize=\footnotesize,
               mathescape=true]{python}
  def Node():
    # gather states of child nodes
    s = gather({0})
    c, h = split(s)
    # pull the external input
    x = pull({0})
    # perform the computation
    i = sigmoid(W|$^{(i)} \times$| x + U|$^{(i)} \times$| h + b|$^{(i)}$|)
    f = sigmoid(W|$^{(f)} \times$| x + U|$^{(f)} \times$| h + b|$^{(f)}$|)
    o = sigmoid(W|$^{(o)} \times$| x + U|$^{(o)} \times$| h + b|$^{(o)}$|)
    u = tanh(W|$^{(u)} \times$| x + U|$^{(u)} \times$| h + b|$^{(u)}$|)
    c = i |$\otimes$| u + f |$\otimes$| c
    h = o |$\otimes$| tanh(c)
    # scatter and push the output
    scatter(concat([c, h], 1))
    push(h)
\end{minted}
\label{fig:seqLSTM}
\end{figure}
\end{comment}

\begin{figure}
\begin{minted}[linenos,
               numbersep=-6pt,
               %gobble=25,
               frame=single,
               framesep=0.3mm,
               fontfamily=tt,,
               escapeinside=||,
               fontsize=\scriptsize,
               mathescape=true]{python}
  def |$\mathcal{F}$|():
    for k in range(N):
      S = gather(k) # gather states of child vertices
      c|$_k$|, h|$_k$| = split(S, 2) # get hidden states c and h 
    x = pull() # pull the first external input x
      
    # specify the computation
    h = |$\sum_{k=0}^{N-1}$|h|$_k$|
    i = sigmoid(W|$^{(i)} \times$| x + U|$^{(i)} \times$| h + b|$^{(i)}$|)
    for k in range(N):
      f|$_k$| = sigmoid(W|$^{(f)} \times$| x + U|$^{(f)} \times$| h|$_k$| + b|$^{(f)}$|)
    o = sigmoid(W|$^{(o)} \times$| x + U|$^{(o)} \times$| h + b|$^{(o)}$|)
    u = tanh(W|$^{(u)} \times$| x + U|$^{(u)} \times$| h + b|$^{(u)}$|)
    c = i |$\otimes$| u + |$\sum_{k=0}^{N-1}$|f|$_k$| |$\otimes$| c|$_k$|
    h = o |$\otimes$| tanh(c)
    
    scatter(concat([c, h], 1)) # scatter c, h to parent vertices
    push(h)                    # push to external connectors
\end{minted}
\vspace{-12pt}
\caption{\small An N-ary child-sum TreeLSTM~\cite{tai2015improved} in Cavs.} %\gn{This is actually a tricky example. How do you batch vertices with different $N$?}}
\label{fig:treeLSTM}
\vspace{-15pt}
\end{figure}

\noindent \textbf{Expressiveness.}
With these four APIs, Cavs can be seen as a middle ground between static and dynamic declaration: In the best case, the model can be easily represented by a single vertex function plus input graphs. While in the worse case scenario, that every sample has a unique input graph while every vertex in the graph has a unique way to interact with its neighboring vertices, Cavs reduces to dynamic declaration that one has to define a vertex function for each vertex of input graphs. However, dynamic NNs in this scenario are very rare and usually not preferred because of the difficulty of design, programming and learning. 

%There might exist other possible design choices, for example, for each node in a Tree-LSTM, customizing the way it interacts with its child nodes, yet this usually results in the model being unable to capture the patterns we intend to learn (not learnable), or inapplicable on various (and potentially arbitrary) input structures (because we then have to customize for each incoming tree, which is apparently impossible).

% (Hao) This does not make sense to me. It does not show how the recurrent structure is connected to the external. It instead confused me....
%|$\mathcal{G}$| = Placeholder();
%Sym external_input = Placeholder();
%NewModel model(graph_structure, external_input);
%Sym loss = model.output().Reduce_sum();

%\subsection{Design and Optimization}
%\label{sec:design}
%In this section, we describe Cavs' system implementation and our proposed optimization strategies derived from Cavs' programming model. 

%The abstraction of vertex (subgraph) for the dynamic neural networks clearly reveals the direction of performance optimizations: the batching of independent vertex and the compilation of the subgraph.For the former, the dependency is different from sample to sample and therefore the batching policy can only be determined in runtime. For the latter, the subgraph is the same for all the samples and therefore its overhead only exists in the initialization.

% \vspace{-12pt}
\subsection{Scheduling}
\vspace{-3pt}
\label{sec:scheduling}
Once users define the vertex function $\mathcal{F}$ and launch the execution, the Cavs scheduler arranges the evaluation of $\mathcal{F}$ over the input graphs.
%following a few policies. %In this section, we first describe the scheduler design, based on which, we further describe the memory management mechanism and the graph optimization strategies.
Given $\mathcal{F}$, Cavs's scheduler follows designed policies to efficiently perform backpropagation for all samples $\{x_i\}_{i=1}^N$ and their associated graphs $\{\mathcal{G}——i\}_{i=1}^N$.

%We next describe how the backpropagation algorithm is executed by Cavs' scheduler over a given sample $x_i$, then introduce two policies that schedule the execution for multiple samples.

\noindent \textbf{Backpropagation.} 
Cavs performs backpropagation~\cite{hinton2006reducing} as follows. For a sample $x_i$ with its input graph $\mathcal{G}_i$, the scheduler starts the forward pass from the input vertices of $\mathcal{G}_i$, and proceeds following the direction indicated by the edges in $\mathcal{G}_i$: at each sub-step, the scheduler figures out the next activated vertex in $\mathcal{G}_i$, and evaluates $\mathcal{F}$ at this vertex following the symbolic programs in $\mathcal{F}$. It then marks this vertex as \emph{evaluated}, and proceeds with the next activated vertex until reaching a terminal vertex (e.g. the loss function). 
A vertex of $\mathcal{G}$ is activated if and only if all its dependent vertices have been evaluated.
%An input vertex of $\mathcal{G}_i$ is activated if and only if $x_i$ was fed as input, while an intermediate vertex is activated if and only if all its dependent vertices have been marked as ``evaluated''. 
The backward pass is continued right after the forward. The scheduler first resets the status of all vertices as \emph{not evaluated}, then scans the graph in a reverse direction, starting from the ending point of the forward pass. It similarly figures out the next activated vertex, but applies another function $\partial \mathcal{F}$, which is the backward function of $\mathcal{F}$ and automatically derived by Cavs via auto-differentiation (\S\ref{sec:autodiff}), until all vertices have been evaluated in backward. %While Cavs inherits the vertex-centric APIs in graph computing and the dataflow graph representations for ML, we note the following key differences in scheduling: at each iteration  
\begin{algorithm}[htp]
\begin{algorithmic}[1]
\footnotesize
\Function{Forward}{$\{x_k\}_{k=1}^K, \{\mathcal{G}_k\}_{k=1}^K, \mathcal{F}$}
%\State Input $\{x_k\}_{k=1}^K$ to the leaf vertices of $\{\mathcal{G}_k\}_{k=1}^K$ respectively, and mark these leaves as evaluated.
\State set task ID $t \leftarrow 0$, task stack $S \leftarrow \emptyset$.
\While{NOT all vertices in $\{\mathcal{G}_k\}_{k=1}^K$ are \emph{evaluated}}
\State figure out all activated vertices in $\{\mathcal{G}_k\}_{k=1}^K$ as a set $V_t$.
\State push $V_t$ into $\mathcal{S}$.
\State evaluate $\mathcal{F}$ on $V_t$: $\mbox{\texttt{GraphExecute}}(V_t, \mathcal{F})$ (see \S\ref{sec:graph_execution}). 
%by calling the graph execution engine: $\mbox{\texttt{GraphEvaluate}}(V_t, \mathcal{F})$ (see \S\ref{sec:graph_execution}). 
\State set the status of all vertices in $V_t$ as \emph{evaluated}.
\State set $t \leftarrow t + 1$.
\EndWhile
\State \Return $\mathcal{S}$.
\EndFunction
\Function{Backward}{$\mathcal{S}, \{\mathcal{G}_k\}_{k=1}^K, \partial \mathcal{F}$}
%\State reset the status of all vertices in $\{\mathcal{G}_k\}_{k=1}^K$ as \emph{not evaluated}.
\State set $t$ as the size of $\mathcal{S}$.
\While{$\mathcal{S}$ is not empty}
\State pop the top element of $\mathcal{S}$ as $V_t$.
\State Evaluate $\partial \mathcal{F}$ on $V_t$: $\mbox{\texttt{GraphExecute}}(V_t, \partial \mathcal{F})$ (\S\ref{sec:graph_execution}).
%\State set the status of all vertices in $V_t$ as evaluated.
\State set $t \leftarrow t - 1$.
\EndWhile
\EndFunction
\end{algorithmic}
\caption{Backpropagation with the batching policy.}
\label{algo:batching_policy}
\vspace{-0.05cm}
\end{algorithm}
%Hao: maybe state some differences between Cavs and traditional graph computing systems.%For the scheduling part, our design of embedding thinking like a vertex in data-flow graph is different from the conventional graph computing and also from data-flow graph. For the former, although both methods activate some vertex to be execute in the next round, our design activate each vertex exactly once in the forward pass, and activate reversely in the backward pass. For the latter, vertex/operator in conventional data-flow graph should be executed exactly once. However, our subgraph instances have to be computed any number of times. The runtime has to record the instance id of the executing and executed instances.
To train a NN to convergence, the above process has to be iterated by the scheduler over all samples $\{x_i\}_{i=1}^N$ and their associated graphs $\{\mathcal{G}_i\}_{i=1}^N$, for many epochs. Instead a sequential execution, Cavs designs a batching policy to perform batched computation, considering the fact that evaluating a set of same arithmetic operations together is significantly faster than the sequential evaluation of each of them.%arrange the execution of $\mathcal{F}$ and $\partial \mathcal{F}$ over $\{x_i\}_{i=1}^N$ and their associated graphs $\{\mathcal{G}_i\}_{i=1}^N$.
%\noindent \textbf{Serial policy}.
%With this policy, the scheduler simply processes one graph at a time, while within each graph it scans the vertex one by one. Despite simplicity, this policy is obviously inefficient as it has to launch as many computational kernels as the number of vertices in $\{\mathcal{G}_i\}_{i=1}^N$ (on some devices, kernel launching may take as much time as computation), and it does not leverage batched computation -- evaluating a set of same arithmetic operations together is significantly faster than the sequential evaluation of each of them.
% \vspace{-0.2cm}
%\vspace{-5pt}

\noindent \textbf{Batching policy}. Given a data batch $\{x_k\}_{k=1}^K \subseteq \{x_i\}_{i=1}^N$ and associated graphs $\{\mathcal{G}_k\}_{k=1}^K$, this policy groups multiple vertices and then performs batched evaluation of $\mathcal{F}$ in order to reduce kernel launches and exploit parallelism.
% Hao: a paragraph to explain the algorithm
Algorithm~\ref{algo:batching_policy} details this policy. 
Specifically, the scheduler divides the forward evaluation of (a batch of) $K$ graphs into multiple steps. At each step $t$, it analyzes $\{\mathcal{G}_k\}_{k=1}^K$ at runtime and determines a set $V_t$ contains all activated vertices in $\{\mathcal{G}_k\}_{k=1}^K$. It then evaluates $\mathcal{F}$ over these vertices by creating a \emph{batching execution task}, with the task ID set to $t$\footnote{Without ambiguity, we use $V_t$ to denote both the set of vertices to be batched together, as well as the batching task itself.}. The task is passed to a graph execution engine, which will further optimize the execution and conduct the actual computation (\S\ref{sec:graph_execution}). 
Meanwhile, the scheduler records the information of this task by pushing $V_t$ into a stack $\mathcal{S}$. At each step of the backward, the scheduler pops out an element $V_t$ from $\mathcal{S}$, creates a corresponded backward batching task -- the execution engine will evaluate the derivative function $\partial \mathcal{F}$ over vertices in $V_t$, until all vertices of $\{\mathcal{G}_k\}_{k=1}^K$ are evaluated.

We note the batching policy plays a similar role as the \emph{dynamic batching} in Fold~\cite{looks2017deep} and DyNet~\cite{neubig2017dynet} in the scenario of dynamic declaration. However, Cavs determines how to batch fully dynamically during runtime using simple breadth-first search with negligible cost (instead of analyzing graphs before every iteration of the execution). We next describe an improved management management strategy based on this batching policy.

%It is worthy to be noted that, batching method in our system is different in design from tf-fold and dynet. Our system is fully dynamic and the dependencies are parsed in the runtime. It is enabled by the four primitives which isolate the subgraph scheduling from specific dependency data. The subgraph instance can be called directly by the scheduler, according to the runtime dependencies. However, tf-fold and dynet have to construct a big data-flow graph which contains all the subgraph instances and the depth is parsed and marked in the graph construction phase.

%\vspace{-6pt}
\subsection{Memory Management}
%\vspace{-2pt}
\label{sec:memory_management}
\begin{figure*}
\centering
\includegraphics[width=0.85\textwidth]{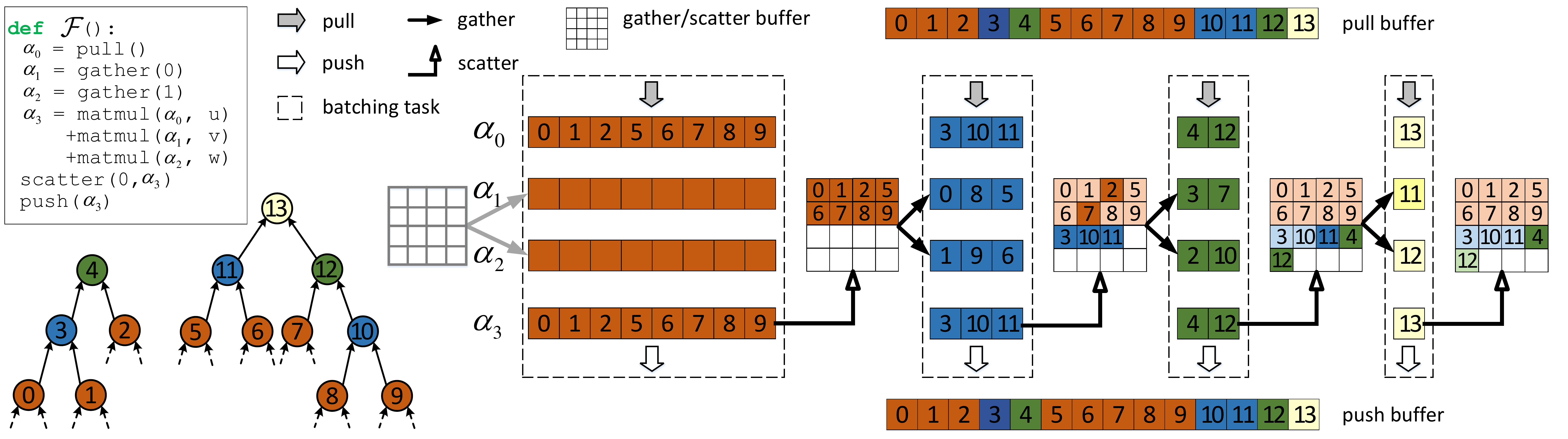}
\vspace{-5pt}
\caption{\small A color or a dashed lined box corresponds to a batching task. The rectangles are memory blocks. The numbers are vertex IDs. Memory blocks in one row belong to a dynamic tensor (e.g. $\alpha_0$ ) and are physically continuous, though we separate them in different boxes.}
%\gn{I didn't understand the ``for presentation purposes'' part, or what $t_0$ is. The lightened colors in the gather/scatter buffer were confusing to me at first. After some thought I realized they were currently in-use arguments.}
\vspace{-10pt}
\label{fig:memory_management}
\end{figure*}
In static declaration~\cite{abadi2016tensorflow,neubig2017dynet}, a symbol in the user program usually corresponds to a \emph{tensor} object, with its shape inferred from the program and batch size specified in advance. The framework usually preallocates a continuous storage on memory for each tensor and fixes it throughout runtime. However, in Cavs, as each batching task is determined at runtime and not visible before execution, its memory management exhibits more complexities.
%\gn{Actually, if it's OK to have the same minibatches for each epoch, we could compile the Cavs graphs before even starting training, right? I think this would be really cool! No need to fix it for this paper though.}
For the batched computation to be efficient, Cavs must guarantee the inputs and intermediate states during the evaluation of $\mathcal{F}$ over a group of runtime-determined vertices coalescing in memory. If we adopt the aforementioned strategy, for each operation in $\mathcal{F}$, Cavs has to index each slice of its input tensor (which may be scattered in different places) and rearrange them as a continuous memory block, which might cause nontrivial overhead.

Cavs proposes a novel data structure \emph{dynamic tensor} to address this challenge (Figure~\ref{fig:dynamic_tensor}). A dynamic tensor is a wrapper of a multi-dimensional array~\cite{abadi2016tensorflow,walt2011numpy} that contains four main members: \texttt{shape}, \texttt{bs}, a pointer \texttt{p} to a chunk of memory, and \texttt{offset}. 
\texttt{shape} is an array of integers representing the specific shape of the tensor excluding the batch dimension. It can be inferred from the user program and set before execution. The batch size \texttt{bs} is implemented as a placeholder in Cavs, with its value dynamically set by the scheduler at runtime at the beginning of a batching task. For each dynamic tensor, Cavs will preallocate a chunk of continuous memory, and point $\texttt{p}$ to its starting address. This memory block is often very large and not fixed-sized -- it can be further extended if needed. To access a dynamic tensor, the execution engine moves \texttt{p} forward with the value specified in \texttt{offset}, and reads/writes number of elements equal to $\mbox{\texttt{bs}} \cdot \prod_i \mbox{\texttt{shape}}[i]$. Therefore, \texttt{bs} together with \texttt{offset} provide a view of the tensor, and the state of the tensor will vary based on their values.
Given a vertex function $\mathcal{F}$, Cavs creates dynamic tensors $\{\alpha_n\}_{n=1}^N$ for each non-parameter symbol $s_n (n = 1, \dots, N)$ in $\mathcal{F}$, and also $\{\nabla \alpha_n\}_{n=1}^N$ as their gradients, while it creates static tensors for model parameters.% i.e. it allocates $2N$ blocks of the memory, points \texttt{p} of those tensors to those memory, then sets their \texttt{offset} to zero and \texttt{shape} by inferring from $\mathcal{F}$, while leaving \texttt{bs} as blank. 
%\shizhen{Cavs treats gradient as dynamic tensors while stores model parameters as normal static tensors.}
\setlength{\columnsep}{5pt}%
\begin{wrapfigure}{r}{0.19\textwidth}
\centering
%\vspace{-5pt}
\begin{subfigure}{0.18\textwidth}
  \centering
  \begin{minted}[%linenos,
               numbersep=0pt,
               %gobble=25,
               frame=single,
               framesep=-1.8mm,
               fontfamily=tt,
               escapeinside=||,
               fontsize=\scriptsize,
               mathescape=true]{cpp}
  struct DynamicTensor {
    vector<int> shape; 
    int bs;
    int offset;
    void* p; };
\end{minted}
\end{subfigure}
\vspace{-10pt}
\caption{\small Dynamic tensor.}
\label{fig:dynamic_tensor}
\vspace{-10pt}
\end{wrapfigure}
Figure~\ref{fig:memory_management} illustrates how the memory is assigned during the forward pass by simply manipulating dynamic tensors.
In particular, in a training iteration, for a batching task $V_t$, the scheduler sets \texttt{bs} of all $\{\alpha_n\}_{n=1}^N$ to $M_t = |V_t|$ (the number of vertices in $V_t$). The execution engine then performs batched evaluation of each expression in $\mathcal{F}$ one by one.
%, starting from the first calls of \texttt{gather}/\texttt{pull}, and ending at \texttt{scatter}/\texttt{push}. 
For an expression $s_l = \mbox{\texttt{op}}(s_r)$\footnote{Without losing generality the user-defined expressions can be arbitrary, e.g. with more than one argument or return values.}, the engine first reads $\alpha_r$ (the dynamic tensor of the RHS symbol $s_r$) by offsetting $\alpha_r.\mbox{\texttt{p}}$ by $\alpha_r.\mbox{\texttt{offset }}$ then reading a block of $M_t \prod_i \alpha_r.\mbox{\texttt{shape}}[i]$ elements, and presents it as a tensor with batch size $M_t$ and other dimensions specified in $\alpha_r.\texttt{shape}$. It then applies batched computational kernels of the operator \texttt{op} over this continuous block, and writes the results to $\alpha_l$ (the dynamic tensor of the LHS symbol $s_l$) on the continuous block in between $[\alpha_l.\mbox{\texttt{p}} + \alpha_l.\mbox{\texttt{offset}}, \alpha_l.\mbox{\texttt{p}} + \alpha_l.\mbox{\texttt{offset}} + M_t \prod_{i} \alpha_l.\mbox{\texttt{shape}}[i]]$. 
%Evaluating $\mathcal{F}$ will visit all dynamic tensors at least once. 
Upon the completion of $V_t$, the scheduler increases the \texttt{offset} of all $\{\alpha_n\}_{n=1}^N$ by $M_t \prod_{i} \alpha_n.\mbox{\texttt{shape}}[i]$, respectively. It then starts the next batching task $V_{t+1}$ until $\mathcal{F}$ has been evaluated at all vertices of $\{\mathcal{G}_k\}_{k=1}^K$. Hence, intermediate results generated in each batching task at forward pass are stored continuously in the dynamic tensors, and their offsets are recorded.

The scheduler then starts the backward pass. It initializes $\nabla \alpha_n.\texttt{offset}$ for each $n$ as $\alpha_n.\texttt{offset}$. Since the backward execution follows an exactly reverse order of the forward pass (Algorithm~\ref{algo:batching_policy}), the intermediate results generated during forward can be easily accessed by decreasing the \texttt{offset} of $\{\alpha_n\}_{n=1}^N$. Specifically, for each batching task $V_t$ popped out from $\mathcal{S}$, the execution engine sets \texttt{bs} of all dynamic tensors to $M_t$, and for each $\alpha_n$ and $\nabla \alpha_n$, decreases their \texttt{offset} by $M_t \prod_i \alpha_n.\mbox{\texttt{shape}}[i]$.
%so as to move to the starting address where the intermediate results were written. 
%The graph execution engine then starts evaluating $\partial \mathcal{F}$ over $V_t$. 
For an expression $\nabla s_r = \mbox{\texttt{grad\_op}}(\nabla s_l, s_l, s_r)$ in $\partial \mathcal{F}$ that corresponds to $s_l = \mbox{\texttt{op}}(s_r)$ in $\mathcal{F}$ (see \S\ref{sec:autodiff}), the engine reads the current states of $\nabla s_l, s_l, s_r$ (which are continuous on memory) and performs batched execution of \texttt{grad\_op}. Different from the forward pass, the gradient result will be added to the current state of $\nabla s_r$ instead of overwriting. 
%As the backward execution follows an exactly reverse order of the forward pass, the engine only needs to move backward the pointers in the reverse order, to  

%\vspace{-6pt}
At the entrance of $\mathcal{F}$, the vertices $\{v_m\}_{m=1}^{M_t}$ in $V_t$ need to interact with its dependent vertices in previous $V_{t-1}$ to \texttt{gather} their outputs as inputs (L3 of Figure~\ref{fig:treeLSTM}), or \texttt{pull} inputs from the external (L5 of Figure~\ref{fig:treeLSTM}). Cavs maintains memory buffers to enable this (Figure~\ref{fig:memory_management}). The memory buffers are key-value stores where the key is the vertex ID $m$ and the value is a tensor slice corresponding to the results of the \texttt{scatter}ed symbol at vertex $v_m$ (batch size is 1). Cavs provides a query function \texttt{IndexBuffer}(\texttt{op}, \texttt{key}) that returns the value in \texttt{op}'s corresponded \texttt{buffer} given a \texttt{key}. During the execution, a \texttt{gather} expression $s_l = \mbox{\texttt{gather}}(\mbox{\texttt{child\_idx}})$ will trigger memory movements: for each $v_m \in V_t$, the scheduler figures out the vertex ID of its child vertex in the input graph with index \texttt{child\_idx}; it then indexes the content in \texttt{gatherBuffer} with the vertex ID as key, and copies and writes it into $\alpha_l$ as a slice. Similarly, at the exit of $\mathcal{F}$, a \texttt{scatter} expression $\mbox{\texttt{scatter}}(s_r)$ will split the current state of $\alpha_r$ into a few slices, and move them to the \texttt{gatherBuffer} for its parent nodes to \texttt{gather}. The \texttt{push}/\texttt{pull} work similarly with \texttt{gather}/\texttt{scatter}, but over the \texttt{pushBuffer} and \texttt{pullBuffer}, respectively, to communicate messages with the external. 
%Hence, at backend, the four proposed APIs are essentially performing memory re-ordering to guarantee memory continuity before batched execution. 

%the scheduler indexes the vertex ID of 
%the execution engine indexes the child vertex with index $child\_idx$ of each vertex $v_m  \in V_t$ using the \texttt{IndexBuffer} function, and copy them and put them as a slice of the dynamic tensor $\alpha_l$. In opposite, at the exit of $\mathcal{F}$, A \texttt{scatter} expression $\mbox{\texttt{scatter}}(s_r)$ will split the corresponded dynamic tensor $\alpha_r$ into a few slices, and copy them to the \texttt{gatherBuffer} for its parent nodes to access. The \texttt{push}/\texttt{pull} APIs work similarly with texttt{gather}/\texttt{scatter}, but over the \texttt{pullBuffer}.
Algorithm~\ref{algo:memory_management} summarizes the memory management process during forward pass. With this strategy, Cavs guarantees memory continuity for any batched computation in $\mathcal{F}$. Compared to dynamic batching in DyNet, Cavs performs memory movement only at the entrance and exit of $\mathcal{F}$, instead of for each expression (operator), it therefore significantly reduces overhead by memory operations (\S\ref{sec:ablation}).

\vspace{-8pt}
\begin{algorithm}[htp]
\begin{algorithmic}[1]
\footnotesize
\Function{Forward}{$\{V_t\}_{t=1}^T, \{\alpha_n\}_{n=1}^N, \mathcal{F}$}
\For{$t  = 1 \rightarrow T$}
\FFor{$n = 1 \rightarrow N$}  $\alpha_n.\mbox{\texttt{bs}} \leftarrow M_t$ \EndFFor
\For{$\mbox{ each expression like } s_l = \mbox{\texttt{op}} (s_r)$ in $\mathcal{F}$}
\If{$\mbox{\texttt{op}} \in \{\mbox{\texttt{gather}}, \mbox{\texttt{pull}}\}$}
\State $C \leftarrow \prod_{i} \alpha_l.\mbox{\texttt{shape}}[i], q \leftarrow \alpha_l.\mbox{\texttt{p}} + \alpha_l.\mbox{\texttt{offset}}$.
\For{$v_m \in V_t (m = 1 \rightarrow M_t) $}
\State $\mbox{\texttt{src}} \leftarrow \mbox{\texttt{IndexBuffer}}(\mbox{\texttt{op}}, m), \mbox{\texttt{dest}} \leftarrow q + (m-1)C$.
\State $\mbox{\texttt{memcpy}}(\mbox{\texttt{dest}}, \mbox{\texttt{src}}, C)$.
\EndFor
\ElsIf{$\mbox{\texttt{op}} \in \{\mbox{\texttt{scatter}}, \mbox{\texttt{push}} \}$}
\State $C \leftarrow \prod_{i} \alpha_r.\mbox{\texttt{shape}}[i], q \leftarrow \alpha_r.\mbox{\texttt{p}} + \alpha_r.\mbox{\texttt{offset}}$.
\For{$v_{m} \in V_t (m = 1 \rightarrow M_t) $}
\State $\mbox{\texttt{dest}} \leftarrow \mbox{\texttt{IndexBuffer}}(\mbox{\texttt{op}}, m), \mbox{\texttt{src}} \leftarrow q + (m-1)C$.
\State $\mbox{\texttt{memcpy}}(\mbox{\texttt{dest}}, \mbox{\texttt{src}}, C)$.
\EndFor
\Else
\State perform batched computation: $\alpha_l = \texttt{op\_kernel} (\alpha_r)$.%, write the (batch) results continuously to $\alpha_l$.
\EndIf
\EndFor
\FFor{$n = 1 \rightarrow N$} $\alpha_n.\texttt{offset} += M_t \prod_{i} \alpha_n.\mbox{\texttt{shape}}[i]$ 
%$\alpha_n \in \{\alpha_n\}_{n=1}^N$
\EndFFor
\EndFor
%\State \Return $\{\alpha_n\}_{n=1}^N$.
\EndFunction
\end{algorithmic}
\caption{Memory management at forward pass.}
\label{algo:memory_management}
\end{algorithm}
\subsection{Auto-differentiation}
%\vspace{-2pt}
\label{sec:autodiff}
%For the ease of programming, the most important feature mainstream ML frameworks provide might be the automatic differentiation. Programmers only define the forward pass then the frameworks generate the backward computation and realize thde training procedure. The well known back-propagation algorithm is adopted for the backward pass.
Cavs by nature supports auto-differentiation. Given a vertex function $\mathcal{F}$ it derives $\partial \mathcal{F}$ following the auto-differentiation rules: for each math expression such as $s_l = \mbox{\texttt{op}}(s_r)$ in $\mathcal{F}$, Cavs generates a corresponded backward expression in $\partial \mathcal{F}$ as $\nabla s_r = \mbox{\texttt{grad\_op}}(\nabla s_l, s_l, s_r)$.
For the four proposed operators, with the memory management strategy described above, we note \texttt{scatter} is the backward operator of \texttt{gather} in the sense that if \texttt{gather} collects inputs from \texttt{gatherBuffer} previously written by \texttt{scatter} at the forward pass, a \texttt{scatter} needs to be performed to write the gradients to the \texttt{gatherBuffer} for its dependent vertices to $\texttt{gather}$ at the backward pass. Hence, for an expression like $s_l = \texttt{gather}(\texttt{child\_idx})$ in $\mathcal{F}$, Cavs will generate a backward expression $\texttt{scatter}(\nabla s_l)$ in $\partial \mathcal{F}$. Similarly, the gradient operator of \texttt{scatter} is \texttt{gather}. The same auto-differentiation rule applies for \texttt{push} and \texttt{pull} as well.
\subsection{Graph Execution Engine}
\label{sec:graph_execution}
\begin{figure}[tb]
%\vspace{-5pt}
\includegraphics[width=0.47\textwidth]{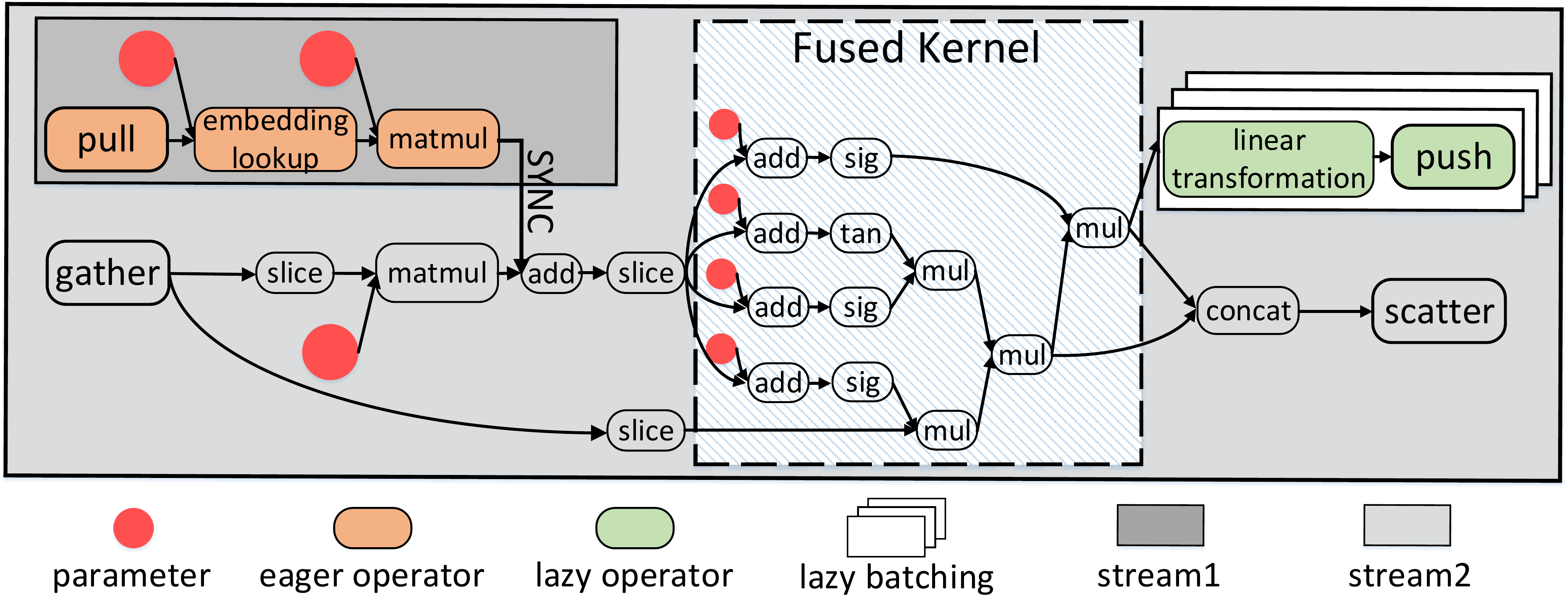}
\vspace{-8pt}
\caption{\small The dataflow graph encoded by $\mathcal{F}$ of Tree-LSTM.}
%\gn{Why is \texttt{slice} not part of the fused kernel?}
\vspace{-12pt}
\label{fig:graph_execution}
\end{figure}
Benefiting from the vertex-centric representation, the vertex function $\mathcal{F}$ essentially defines a (small) static dataflow graph that is open to various graph execution optimization techniques (which might not be the case in dynamic declaration). We next discuss three proposed optimizations on Cavs' execution engine for improved performance.
%how we optimize Cav's graph execuction engine to e
%that will be repeatedly evaluated by the graph execution engine at each vertex of input graphs. 
%While the scheduler speeds up the evaluation by parallelizing the evaluation of $\mathcal{F}$ at multiple vertices, there are still opportunities that the execution of $\mathcal{F}$ within a given batching task $V_t$ can be further optimized.
%Almost all the performance optimizations on modern processors (multi/many-core, GPU, etc) can be regarded as a compromise to the computer architecture. 
%In this section, we discuss Cavs' graph execution engine. 
%we propose three major optimization techniques: \emph{lazy batching}, \emph{streaming}, and \emph{kernel fusion}. We show that benefiting from Cavs' system architecture, these strategies can be easily incorporated and help further exploit potential parallelism  during the evaluation of a single batching task.
%enlarging the parallelism, better usage of faster memory in the memory hierarchy, and some specific features of certain architectures. 

\noindent \textbf{Lazy batching and streaming\footnote{Streaming is a borrowed terminology from CUDA programming which means executing different commands concurrently or out of order with respect to each other on different GPU streams. As Cavs' optimization strategies are agnostic to the low-level hardware, we use streaming interchangeably with multi-threading if the underlying computing hardware is CPU.}.}
In addition to batched execution of $\mathcal{F}$, the lazy batching and streaming explore potential parallelism for a certain group of finer-grained operators in $\mathcal{F}$ or $\partial \mathcal{F}$ called lazy and eager operators.
%In particular, when the user defines the vertex function $\mathcal{F}$, Cavs will analyze its structure, and figure out a group of lazy and eager operators in $\mathcal{F}$ and $\partial \mathcal{F}$, with them defined as follows.
\begin{definition}
\vspace{-5pt}
An operator in $\mathcal{F}$ ($\partial \mathcal{F}$) is a \emph{lazy operator} if at the forward (backward) pass, for $\forall v \in \mathcal{G}, \forall \mathcal{G} \in \{\mathcal{G}_k\}_{k=1}^K$, the evaluation of $\mathcal{F}$ ($\partial \mathcal{F}$) at any parent (dependent) vertex of $v$ does not rely on the evaluation of $\mathcal{F}$ at $v$. It is an \emph{eager operator} if the evaluation at $v$ does not rely on the evaluation of $\mathcal{F}$ ($\partial \mathcal{F}$) at any dependent (parent) vertices of $v$.
\vspace{-5pt}
\end{definition}
In Cavs, figuring out eager and lazy operators in $\mathcal{F}$ and $\partial \mathcal{F}$ is straightforward given the following proposition:
\begin{proposition}
\vspace{-5pt}
Denote $\mathcal{D}_{\mathcal{F}}$ ($\mathcal{D}_{\partial \mathcal{F}}$) as the dataflow graph encoded by $\mathcal{F}$ ($\partial \mathcal{F}$), and $g, s \in \mathcal{D}_{\mathcal{F}}$ ($\mathcal{D}_{\partial \mathcal{F}}$) as corresponded nodes of the \texttt{gather} and \texttt{scatter} operator, respectively. A node that has $g$ as its dependent and is not on any path from $g$ to $s$ is a lazy operator. A node that has $s$ as its ancestor and is not on any path from $g$ to $s$ is an eager operator.
\vspace{-5pt}
\end{proposition}
Figure~\ref{fig:graph_execution} illustrates a dataflow graph of the vertex function of Tree-LSTM, with eager and lazy operators colored. A property of them is that their evaluation is not fully subject to the dependency reflected by the input graph $\mathcal{G}$. For instance, the \texttt{pull} operator in Figure~\ref{fig:graph_execution} is eager and can be executed in prior -- even before $\mathcal{F}$ has been evaluated at the vertices \texttt{gather} tries to interact with; the \texttt{push} operator is lazy, so we can defer its execution without impacting the evaluation of $\mathcal{F}$ at parent vertices. Similarly, in $\partial \mathcal{F}$, the gradient derivation for model parameters are mostly lazy -- their execution can be deferred as long as the gradients of hidden states are derived and propagated in time. 
Cavs leverages this property and proposes the lazy batching strategy. It defers the execution of all lazy operators in $\mathcal{F}$ and $\partial \mathcal{F}$ until all batching tasks $\{V_t\}_{t=1}^T$ has finished. It then performs a batched execution of these lazy operators over all vertices of $\{\mathcal{G}_k\}_{k=1}^K$. These operators includes, but is not limited to, the \texttt{push} operator that are doing memory copy, and the math operators for computing gradients of the model parameters. Lazy batching helps exploit more parallelism for the execution of lazy operators and significantly reduces kernel launches. Empirically lazy batching brings $20\%$ overall improvement (see \S\ref{sec:ablation}). 

On the other hand, we are unable to ``eagerly'' batch eager operators, as their execution over some vertices relies on knowing the detailed memory location of all intermediate results in advance, a condition which is not satisfied in Cavs where memory is dynamically assigned. To leverage the exhibited parallelization opportunity between eager operators and the operators on the path from \texttt{gather} to \texttt{scatter} (Figure~\ref{fig:graph_execution}), Cavs proposes a streaming strategy that pipelines the execution of these two groups of operators. It allocates two streams, and puts the eager operators on one stream, and the rest (excluding lazy operators) on the other. Hence, independent operators in two streams run in parallel, while for those operators that depend on an eager operator, this dependency is respected by synchronization barriers (see Figure~\ref{fig:graph_execution}). It is also possible to parallelize independent paths from $g$ to $s$ on the graph, but we find it does not yield further improvement.

\noindent \textbf{Automatic kernel fusion.}
Since Cavs abstracts out a static dataflow graph encoded by $\mathcal{F}$ that will be dynamically evaluated elsewhere, we can replace the original $\mathcal{F}$ with an optimized one that runs more efficiently, as long as it accepts the same input and produces the same output. %In Cavs, the graph execution engine is alternatively working on a new graph generated from the original one based on kernel fusion mechanisms~\cite{gysi2015stella}.

% automatically generate new kernels using NVRTC
% replace the old subgraph with a new opeartor
Particularly, given $\mathcal{F}$, before execution, Cavs will run a \emph{fusion detector}~\cite{gysi2015stella} to scan its corresponded dataflow graph  and report all \emph{fuse-able} subgraphs therein, i.e. all nodes in a fuse-able subgraph can be fused as a single operator that behaves equivalently but takes less execution time (e.g. with fewer kernel launches and I/O, or faster computation). Currently, we only detect groups of directly linked elementwise operators, such as $+, -, \times, \div, \texttt{tanh}, \texttt{sigmoid}$, as shown in Figure~\ref{fig:graph_execution}, and we use a simple union-find algorithm to detect the largest possible fuse-able subgraphs. Given a fuse-able subgraph, Cavs adopts de facto automatic code generation techniques~\cite{quinlan2000rose, dave2009cetus, ragan2013halide, nvrtc} to generate lower-level kernel codes as an implementation for it. Replacing the original fuse-able subgraphs with fused operators during execution is beneficial in many aspects: (1) it reduces the number of kernel launches; (2) on some devices such as GPUs, kernel fusion transform device memory access into faster device registers access. We empirically report another $20\%$ improvement with automatic kernel fusion (\S\ref{sec:ablation}).
\section{Implementation}
\label{sec:implementation}
%Describe our system implementation and architectures here.
We implement Cavs as a pluggable C++ library that can be integrated with existing DL frameworks to provide or enhance their support for dynamic NNs. We next briefly discuss implementation details. For clarity, we assume the host framework is composed of three major layers (which is the case for most popular frameworks~\cite{Bergstra:2011:NIPSW,abadi2016tensorflow,neubig2017dynet}): (1) a frontend that provides device-agnostic symbolic programming interface; (2) an intermediate layer that implements the core execution logic; (3) a backend with device-specific kernels for all provided operators. %We next brief what additional implementations are needed at each layer of the host framework.

\noindent \textbf{Frontend.}
%The frontend layer of a DL framework usually provides high-level symbolic interface for a collection of primitive operators which the programmers directly interact with. Users can assemble complex operations by combining primitive ones. 
Cavs provides a base class \texttt{GraphSupport} in addition to conventional operators and the four proposed APIs (\S\ref{sec:api}). Users are required to instantiate it by providing a symbolic vertex function $\mathcal{F}$ -- therefore an instantiation of \texttt{GraphSupport} represents a single dynamic structure. %To construct a dynamic network, an instantiation of \texttt{GraphSupport} maps to a single dynamic NN. To construct a dynamic network, 
To construct more complex structures (e.g. encoder-decoder LSTM~\cite{sutskever2014sequence}, LRCN~\cite{donahue2015long})), users employ \texttt{push} and $\texttt{pull}$ to connect this dynamic structure to external structures.

%instana DynamicGraph can be connected with another DynamicGraph (e.g. encoder-decoder LSTM~\cite{}), or with a traditional static dataflow graph (e.g. the long-term recurrent convolutional networks~\cite{}).

%Cavs impl introduces a \textit{GraphSupport} base class in which there is a virtual node function.
%Developers need to inherit the GraphSupport class and define override the \textit{node} function to define their model. The constructor of the GraphSupport class needs the symbol of previous layer and the output function of GraphSupport class returns a symbol, which can be used as input of the next layer.
\noindent \textbf{Intermediate Layer.}
At the intermediate layer, Cavs will create a unique scope~\cite{abadi2016tensorflow}, and generates a small dataflow graph for each instantiation of \texttt{GraphSupport}, connecting them appropriately with other parts of the model according to user programs. 
Cavs implements its core runtime logic at this layer, i.e. the scheduler, the memory management, and the graph execution engine, etc. During execution, the execution engine first analyzes the received dataflow graphs and incorporates described optimization in \S\ref{sec:graph_execution}. The scheduler then instructs the system to read training samples and their associates graphs (e.g. adjacency matrices). It starts training by submitting batching tasks to the execution engine and assigning memory accordingly.

\noindent \textbf{Backend.}
Following common practice~\cite{abadi2016tensorflow,neubig2017dynet,caffe2}, Cavs puts device-specific kernel implementations for each supported operator at this layer. Each operator implementation is a function that takes as inputs static tensors and produces static tensors as outputs -- therefore the higher-layer logic, i.e. how the computation is scheduled or how the memory is assigned are invincible to this layer.
Cavs will reuse the native operator implementations from the host framework, while it provides optimized implementations for the four proposed primitives (\texttt{gather}, \texttt{scatter}, \texttt{pull}, \texttt{push}). Specifically, \texttt{gather} and \texttt{pull} index different slices of a tensor and puts them together continuously on memory; \texttt{scatter} and \texttt{push} by contrast splits a tensor along its batch dimension, and copy different slices to different places. Cavs implements a customized memcpy kernel for there four operators so that copying multiple slices from (or to) different places is performed within one kernel, to further reduce kernel launches.

\section{Evaluation}
\label{sec:evaluation}
In this section, we evaluate Cavs on training different NNs across multiple datasets, obtaining the following major findings: (1) Cavs has little overhead: when training static NNs that can be by nature batched, Cavs demonstrates equal performance with other DL systems. On several NNs with notably difficult-to-batch structures, Cavs outperforms all existing frameworks by a large margin; (2) We confirm the graph construction overhead is substantial in both Fold~\cite{looks2017deep} and dynamic declaration~\cite{neubig2017dynet}, while Cavs bypasses it by loading input graphs through I/O; (3) We verify the effectiveness of our proposed design and optimization via ablation studies, and discuss Cavs' advantages over other state-of-the-art DL systems for dynamic dataflow graphs.
% (4) We conduct a user study and show that Cavs' interface is cognitively simpler to ML programmers than Fold.
%It reveals the following results: (1) On recurrent neural network with fixed length, the vertex-centric Cavs introduces little overhead compared with static fully-unrolled dataflow graph. (2) On recurrent neural network with variable length, Cavs is more efficient compared with dynamic unrolled framework(tensorflow and dynet). (3) On recursive neural network, Cavs is both efficient in dependency-graph processing and computation than state-of-the-art batching framework(TF-Fold and dynet auto-batching). As our system enables faster training without changing the models, we only measure the performance in this section.

\noindent \textbf{Environment.} We perform all experiments in this paper on a single machine with an NVIDIA Titan X (GM200) GPU, a 16-core (32 threads) CPU, and CUDA toolkit 8.0 and cuDNN v6 installed. As modern DL models are mostly trained using GPUs, we focus our evaluation on GPUs, but note Cavs' design and implementation do not reply on a specific type of device. We borrow the implementations of most mathematical operators from TensorFlow v1.2, while we implement the four proposed operators and other system modules by ourselves. We mainly compare Cavs to TensorFlow v1.2~\cite{abadi2016tensorflow} with XLA~\cite{xla} and its variant Fold~\cite{looks2017deep}, as well as DyNet v2.0~\cite{neubig2017dynet} with autobatching~\cite{neubig2017fly}, as they have reported better performance than other frameworks~\cite{pytorch,chainer} on dynamic NNs. We focus on metrics for system performance, e.g. the average time to scan one epoch of data. Cavs produces exactly the same numerical results with other frameworks, hence the same per-epoch convergence\footnote{The code of Cavs will be released along with a the next major release of the DyNet project: \url{http://dynet.io/}.}. 

\noindent \textbf{Models and dataset.} We experiment on the following models with increasing difficulty to batch:
(a) \texttt{Fixed-LSTM} language model (LM): a static sequence LSTM with fixed steps for language modeling~\cite{sundermeyer2012lstm,sutskever2014sequence,zaremba2014recurrent}. We train it using the PTB dataset~\cite{ptb} that contains over 10K different words. We set the number of steps as 64, i.e. at each iteration of training, the model takes a 64-word sentence from the training corpus, and predicts the next word of each word therein. Obviously, the computation can be by nature batched easily, as each sentence has exactly the same size.
(b) \texttt{Var-LSTM} LM: that accepts variable-length inputs. At each iteration the model takes a batch of natural sentences with different length from PTB, and predicts the next words; 
(c) \texttt{Tree-FC}: the benchmarking model used in~\cite{looks2017deep} with a single fully-connected layer as its cell function. Following the same setting in~\cite{looks2017deep}, we train it over synthetic samples generated by their code~\cite{treefc-code} -- each sample is associated with a complete binary tree with 256 leaves (therefore 511 vertices per graph);
(d) \texttt{Tree-LSTM}: a family of dynamic NNs widely adopted for text analysis~\cite{liang2016semantic,vinyals2015grammar}. We implement the binary child-sum Tree-LSTM model in~\cite{tai2015improved}, and train it as a sentiment classifier using Stanford sentiment treebank (SST) dataset~\cite{socher2013recursive}, which contains 8544 training sentences in which the longest sentence has 54 words. Each sentence is associated with a human annotated grammar tree.
%This model is usually trained in other frameworks with batch size 1~\cite{tai2015improved,neubig2017dynet,looks2017deep}.
%The tree is an arbitrary binary tree where each word in the sentence maps to one of its leaves.
%sentiment classifier~\cite{tai2015improved} on the Stanford sentiment treeBank (SST) dataset~\cite{socher2013recursive}. 
 %The LSTM unit will be applied following the tree structure from its leaves to the root, in order to predict a sentiment for the whole sentence. 
%which is usually trained by ML researchers with single-instance backpropagation~\cite{neubig2017dynet,looks2017deep}.

%We evaluate Cavs on multiple NN structures and standard dataset, ranging from easy-to-batch NNs to difficult-to-batch ones, which we will report in the following experiment sections. 

\begin{figure*}[tph]
\centering 
\includegraphics[width=0.95\textwidth]{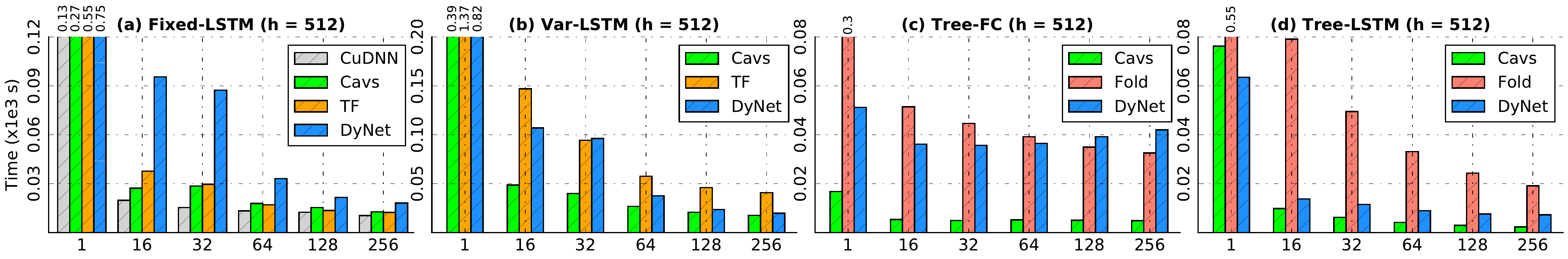} 
\hspace{5pt} \includegraphics[width=0.95\textwidth]{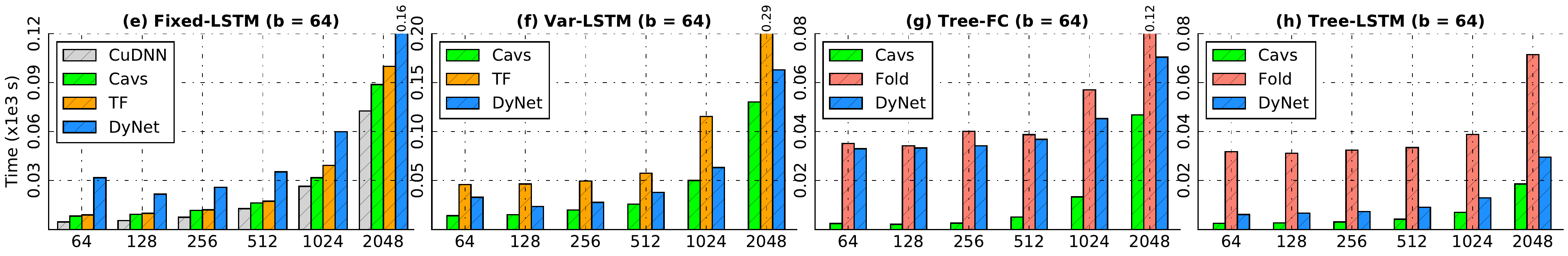}
\vspace{-5pt}
\caption{\small Comparing five systems in terms of the averaged time to finish one epoch of training (lower is better) on four models: \texttt{Fixed-LSTM}, \texttt{Var-LSTM}, \texttt{Tree-FC} and \texttt{Tree-LSTM}. In (a)-(d) we fix the hidden size $h$ and vary the batch size $bs$, while in (e)-(h) we fix $bs$ and vary $h$.}
\vspace{-10pt}
\label{fig:overall_performance}
\end{figure*}

%\vspace{-5pt}
\subsection{Overall Performance}
\label{sec:overall_performance}
%We next evaluate the overall performance of Cavs compared to other state-of-the-art DL systems. 
We first verify the viability of our design on the easiest-to-batch case: \texttt{Fixed-LSTM} language model. We compare Cavs to the following three strong baselines: (1) \texttt{CuDNN}~\cite{chetlur2014cudnn}: a CuDNN-based fixed-step sequence LSTM, which is highly optimized by NVIDIA using handcrafted kernels and stands as the best performed implementation on NVIDIA GPUs; (2) \texttt{TF}: the official implementation of \texttt{Fixed-LSTM} LM in TensorFlow repository~\cite{TF-fixedLSTM} based on static declaration; (3) \texttt{DyNet}: we implement a 64-step LSTM in DyNet based on dynamic declaration -- we declare a dataflow graph per sample, and train with the autobatching~\cite{neubig2017fly} enabled; (4) \texttt{Cavs} with batching policy, and all input samples have a same input graph -- a 64-node chain. We train the model to converge, and report the average time per epoch in Figure~\ref{fig:overall_performance}(a)(e), where in (a) we fix the hidden size $h$ of the LSTM unit as 512 and vary the batch size $bs$, and in (e) we fix $bs = 64$ and vary $h$. Empirically, CuDNN performs best in all cases, but note it is highly inflexible.
%, echoing our statement that the specialized implementation by NVIDIA upper bounds the performance on GPUs, 
Cavs performs slightly better than \texttt{TF} in various settings, verifying that our system has little overhead dealing with fully static graphs, though it is specialized for dynamic ones. We also conclude from Figure~\ref{fig:overall_performance} that batching is essential for GPU-based DL: $bs = 128$ is nearly one order of magnitude faster than $bs = 1$ regardless of used frameworks. For Cavs, the batching policy is 1.7x, 3.8x, 7.0x, 12x, 15x, 25x, 36x faster than the serial policy at $bs = 2, 4, 8, 16, 32, 64, 128$, respectively.
%For completeness, we also report the performance speedup with single-instance training (serial policy): at $bs = 2, 4, 8, 16, 32, 64, 128$ : 1.7x, 3.8x, 7.0x, 12x, 15x, 25x, 36x.
% why dynet is the worst.

Next, we experiment with \texttt{Var-LSTM}, the most commonly used RNN for variable-length sequences. 
We compare the following three implementations (CuDNN-based LSTM cannot handle variable-length inputs): (1) \texttt{TF}: an official TensorFlow implementation based on the dynamic unroll approach described in \S\ref{sec:programming_dynamic}; 
%instead of constructing a single static graph, to handle variable-length inputs, TensorFlow uses its control flow functionality and a bucketing strategy -- at each iteration of the training, the \emph{tf\_while} operator is used to activate a LSTM graph with a time step equal to the length of the longest sentence in the batch. It then appends null words at the end of each sentence in the batch so that all sentences have the same length with the longest sentence, therefore, batch computation can be performed; 
(2) \texttt{DyNet}: an official implementation from DyNet benchmark repository based on dynamic declaration~\cite{DyNet-VarLSTM}; (3) \texttt{Cavs}: where each input sentence is associated with a chain graph that has number of vertices equal to the number of words. 
We vary $h$ and $bs$, and report the results in Figure~\ref{fig:overall_performance}(b)(f), respectively. Although all three systems perform batched computation in different ways, \texttt{Cavs} is constantly 2-3 times faster than \texttt{TF}, and outperforms \texttt{DyNet} by a large margin. Compared to \texttt{TF}, \texttt{Cavs} saves computational resources. \texttt{TF} dynamically unrolls the LSTM unit according to the longest sentence in the current batch, but it cannot prevent unnecessary computation for those sentences that are shorter than the longest one.

We then turn to \texttt{Tree-FC}, a dynamic model for benchmarking.
Since vanilla TensorFlow is unable to batch its computation, we compare \texttt{Cavs} to (1) \texttt{DyNet} and (2) \texttt{Fold}, a specialized library built upon TensorFlow for dynamic NNs, with a depth-based dynamic batching strategy. To enable the batching, it however needs to preprocess the input graphs, translate them into intermediate representations and pass them to lower-level TensorFlow control flow engine for execution.
We report the results in Figure~\ref{fig:overall_performance}(c)(g) with varying $bs$ and $h$, respectively. For all systems, we allocate a single CPU thread for graph preprocessing or construction. \texttt{Cavs} shows at least an order of magnitude speedups than \texttt{Fold} and \texttt{DyNet} at ($h \le 512$). Because the size of the synthetic trees is large, one major advantage of \texttt{Cavs} over them is the alleviation of substantial graph preprocessing/construction overhead. With a single CPU thread, \texttt{Fold} takes even more time on graph preprocessing than computation (\S\ref{sec:ablation}). 

Finally, we compare three frameworks on \texttt{Tree-LSTM} in Figure~\ref{fig:overall_performance}(d)(h): \texttt{Cavs} is 8-10x faster than \texttt{Fold}, and consistently outperforms \texttt{DyNet}. One difference in this experiment is that we allocate as many CPU threads as possible (32 on our machine) to accelerate graph preprocessing for \texttt{Fold}, otherwise it will take much longer time. 
Further, we note \texttt{DyNet} performs much better here than on \texttt{Tree-FC}, as the size of the input graphs in SST (maximally 52 leaves) is much smaller than the synthetic ones (256 leaves each) in \texttt{Tree-FC} experiments. We observe \texttt{DyNet} needs more time on graph construction for large input graphs, and \texttt{DyNet}'s dynamic batching is less effective on larger input graphs, as it has to perform frequent memory checks to support its dynamic batching, which we will discuss in \S\ref{sec:ablation}.

\subsection{Graph Construction and Computation}
\label{sec:graph_construction}
\begin{figure}
\centering
\includegraphics[width=0.47\textwidth]{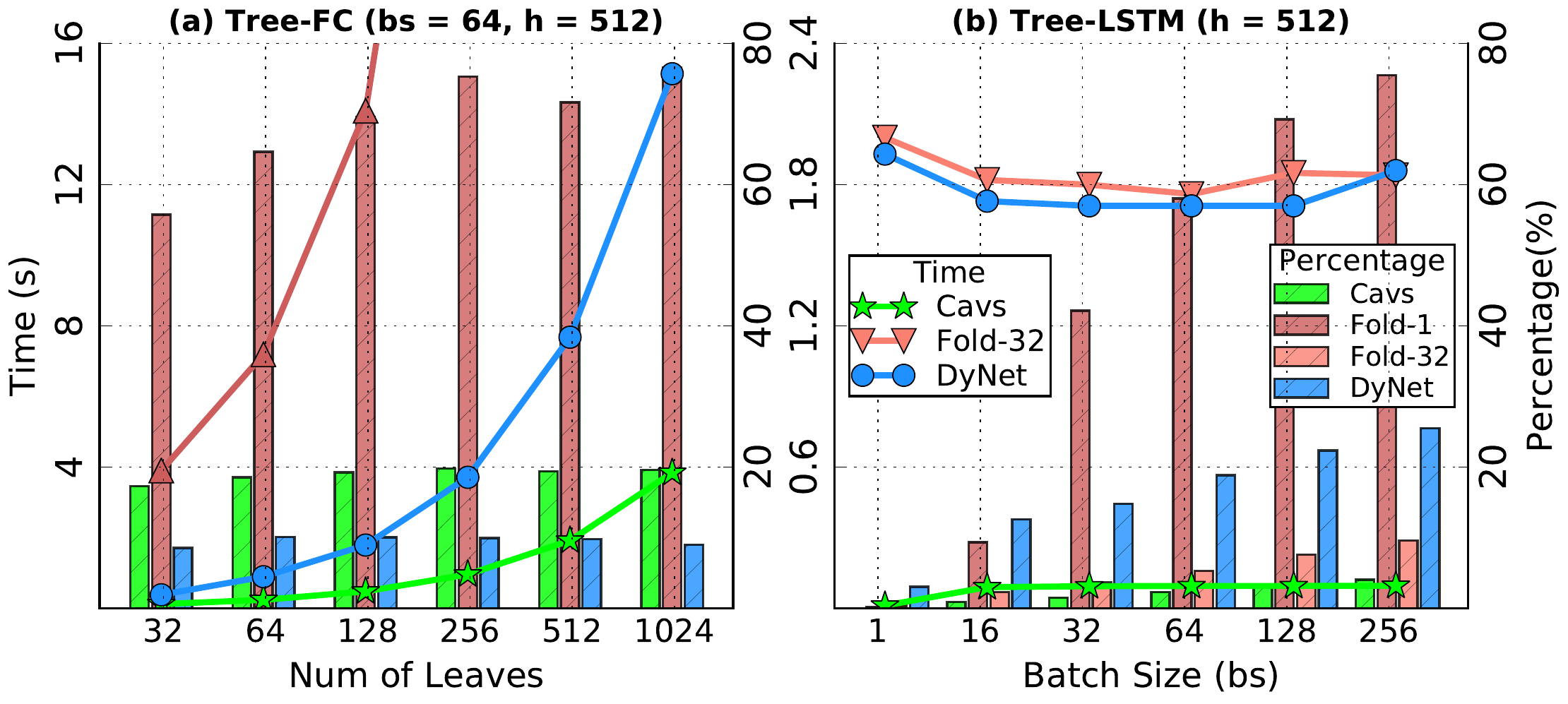}
\vspace{-5pt}
\caption{\small The averaged graph construction overhead per epoch when training (a) \texttt{Tree-FC} with different size of input graphs (b) \texttt{Tree-LSTM} with different batch size. The curves show absolute time in second (left $y$-axis), and the bar graphs show its percentage of the overall time (right $y$-axis).}
\vspace{-10pt}
\label{fig:graph_construction}
\end{figure}

In this section, we investigate the graph construction overhead in Fold and DyNet. To batch computation of different graphs, Fold analyzes the input graphs to recognize batch-able dynamic operations, then translates them into intermediate instructions, with which, TensorFlow generates appropriate control flow graphs for evaluation -- we will treat the overhead caused in both steps as Fold's graph construction overhead. DyNet, as a typical dynamic declaration framework, has to construct as many dataflow graphs as the number of samples. Though DyNet has optimized its graph construction to make it lightweight, the overhead still grows with the training set and the size of input graphs. By contrast, Cavs takes constant time to construct a small dataflow graph encoded by $\mathcal{F}$, then reads input graphs through I/O. To quantify the overhead, we separate the graph construction from computation, and visualize in Figure~\ref{fig:graph_construction}(a) how it changes with the average number of leaves (graph size) of input graphs on training \texttt{Tree-FC}, with fixed $bs = 64, h = 512$. We compare (1) \texttt{Cavs} (2) \texttt{Fold-1} which is Fold with one graph processing thread and (3) \texttt{DyNet}. We plot for one epoch, both the (averaged) absolute time for graph construction and it percentage of the overall time. Clearly we find that all three systems take increasingly more time when the size of the input graphs grows, but Cavs, which loads graphs through I/O, causes the least overhead at all settings. In terms of the relative time, \texttt{Fold} unfortunately wastes 50\% at 32 leaves, and 80\% when the tree has 1024 leaves, while \texttt{DyNet} and \texttt{Cavs} take only $10\%$ and $20\%$, respectively.

We also wonder how the overhead is related with batch size when there is fixed computational workload. We report in Figure~\ref{fig:graph_construction}(b) the same metrics when training \texttt{Tree-LSTM} with varying $bs$. We add another baseline \texttt{Fold-32} with 32 threads for Fold's graph preprocessing. As \texttt{Fold-1} takes much longer time than others, we report its time at $bs = 1, 16, 32, 64, 128, 256$ here (instead of showing in Figure~\ref{fig:graph_construction}): 1.1, 7.14, 31.35, 40.1, 46.13, 48.77. Except $bs = 1$,  all three systems (except \texttt{Fold-1}) take almost constant time for graph construction in one epoch, regardless of $bs$, while \texttt{Fold-32} and \texttt{DyNet} take similar time, but \texttt{Cavs} takes 20x less. Nevertheless, at the percentage scale, increasing $bs$ makes this overhead more prominent, because larger batch size yields improved computational efficiency, therefore less time to finish one epoch. This, from one perspective, reflects that the graph construction is a main obstacle that grows with the number of training samples and prevents the efficient training of dynamic NNs in existing frameworks, while Cavs successfully overcomes this barrier through its design.

Apart from the graph construction we report in Table~\ref{tab:computation_only} the computation-only time -- \texttt{Cavs} demonstrates maximally 5.4x/9.7x and 7.2x/2.4x speedups over \texttt{Fold}/\texttt{DyNet} on \texttt{Tree-FC} and \texttt{Tree-LSTM}, respectively. Besides less system overhead, the advantages stem from two main sources: an optimized graph execution engine, and a better-suited memory management strategy, which we investigate next.

\begin{table}[!ht]
\scriptsize
\centering
\begin{tabular}{ |K{0.42cm}|K{1.6cm}|K{1.05cm}||K{0.23cm}|K{1.6cm}|K{1.05cm}|} 
\hline
\# leaves & Comp. time (s) & Speedup & $bs$ & Comp. time (s) & Speedup\\

%& \cline{4-6} \\
\hline
%\hline
%0.57472, 1.05728, 2.01856, 3.88992, 7.961600000000001, 15.756
%3.0976, 3.93472, 6.15424, 10.6368, 18.4832, 32.4352
%4.10496, 7.974399999999999, 16.0, 33.664, 70.6048, 152.96
32   &  \textbf{0.58} /  3.1 / 4.1  & \textbf{5.4x} / 7.1x & 1  & 76.2  / 550  / \textbf{61.6}    & \textbf{7.2x} / 0.8x\\
64   &  \textbf{1.1}  /  3.9 / 8.0  & 3.7x / 7.5x & 16  & \textbf{9.80}  / 69 / 12    & 7.0x / 1.2x \\
128  &  \textbf{2.0}  /  6.2 / 16.0  & 3.0x / 7.9x & 32  & \textbf{6.15}  / 43 / 9.9    & 7.0x / 1.6x\\
256  &  \textbf{3.9}  / 10.6  / 33.7  & 2.7x / 8.7x & 64  & \textbf{4.1}  / 29 / 7.4    & \textbf{7.2x} / 1.8x\\
512  &  \textbf{8.0}  / 18.5  / 70.6  & 2.3x / 8.9x & 128 & \textbf{2.9}  / 20.5 / 5.9    & 7.1x / 2.0x \\
1024 &  \textbf{15.8}  / 32.4  / 153   & 2.1x / \textbf{9.7x} & 256 & \textbf{2.3}  / 15.8 / 5.4    & 7.0x / \textbf{2.4x} \\
\hline
%\hline
%$bs$ & Computation time & Speedup  \\
%\hline
%\hline
%%76.24, 9.8, 6.15, 4.05, 2.9, 2.27
%%549.6, 69, 43, 29, 20.5, 15.8
%%61.6, 12, 9.85, 7.36, 5.94, 5.44
%1   & 76.2  / 550  / \textbf{61.6}    & \textbf{7.2x} / 0.8x \\
%16  & \textbf{9.80}  / 69.0 / 12.0    & 7.0x / 1.2x \\
%32  & \textbf{6.15}  / 43.0 / 9.85    & 7.0x / 1.6x  \\
%64  & \textbf{4.05}  / 29.0 / 7.36    & \textbf{7.2x} / 1.8x  \\
%128 & \textbf{2.90}  / 20.5 / 5.94    & 7.1x / 2.0x  \\
%256 & \textbf{2.27}  / 15.8 / 5.44    & 7.0x / \textbf{2.4x}  \\
%\hline
\end{tabular}
\vspace{-5pt}
\caption{\small The computation time in second (\texttt{Cavs}/\texttt{Fold}/\texttt{DyNet}) and the speedup (\texttt{Cavs} vs \texttt{Fold}/\texttt{DyNet}) for training one epoch on \texttt{Tree-FC} with varying size of the input trees (left part), and on \texttt{Tree-LSTM} with varying batch size (right part).}
\vspace{-10pt}
\label{tab:computation_only}
\end{table}

%\vspace{-7pt}
\subsection{Ablation Studies}
\begin{figure}[bt]
\centering
\vspace{-5pt}
\includegraphics[width=0.47\textwidth]{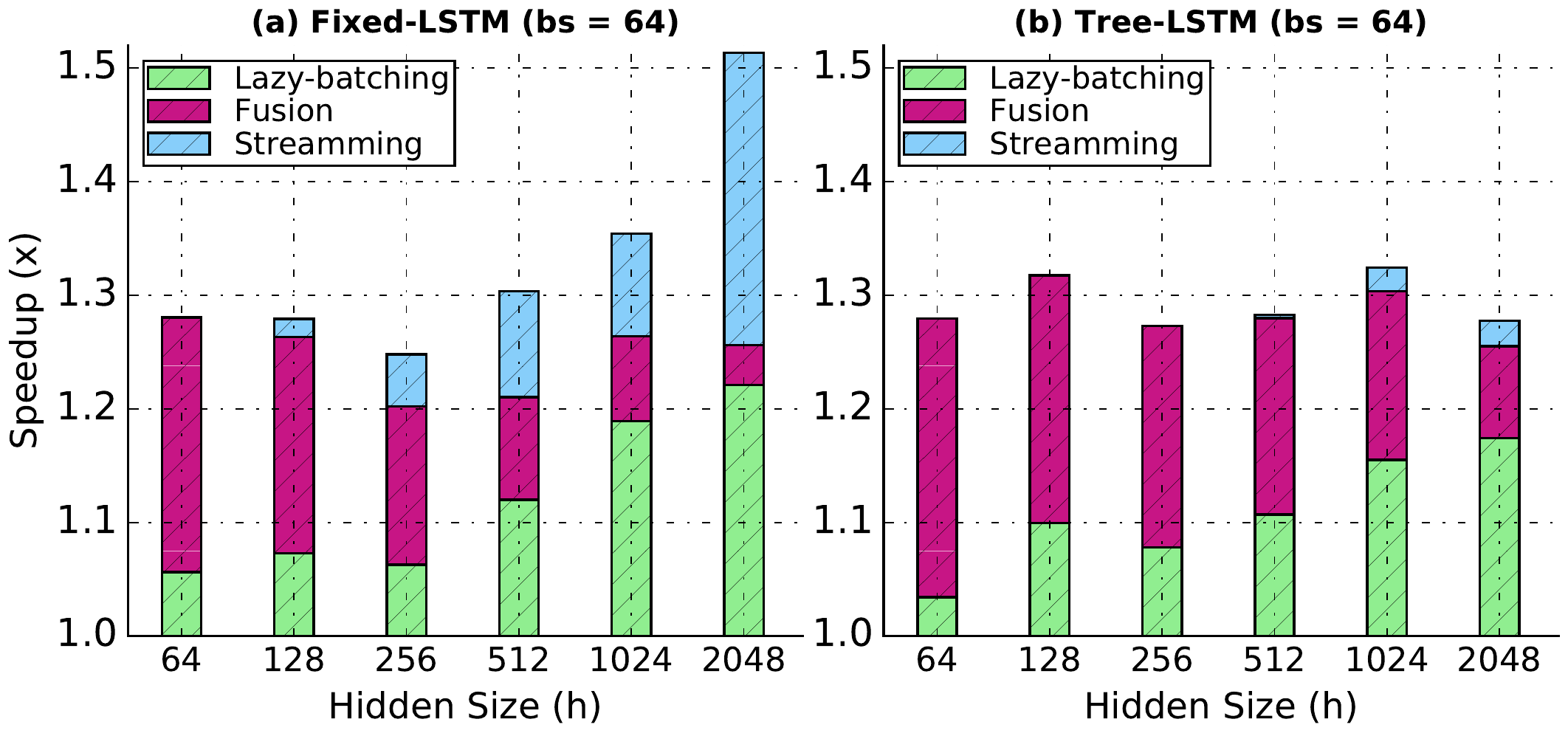}
\vspace{-5pt}
\caption{\small Improvement of each optimization strategy on execution engine over a baseline configuration (speedup = 1).}
\label{fig:ablation}
\vspace{-12pt}
\end{figure}

%\vspace{-3pt}
\label{sec:ablation}
\noindent \textbf{Graph Execution Engine.} To reveal how much each optimization in \S\ref{sec:graph_execution} contributes to the final performance, we disable lazy batching, fusion and streaming in Cavs and set this configuration as a baseline (speedup = 1). We then turn on one optimization at a time and record how much speedup it brings. We train \texttt{Fixed-LSTM} and \texttt{Tree-LSTM}, and report the averaged speedups one computation-only time in one epoch over the baseline configuration in Figure~\ref{fig:ablation}, with $bs=64$ but varying $h$. Lazy batching and fusion consistently deliver nontrivial improvement -- lazy batching is more beneficial with a larger $h$ while fusion is more effective at smaller $h$, which are expected: lazy batching mainly parallelizes matrix-wise operations (e.g. \texttt{matmul}) commonly with $O(h^2)$ our higher complexity, while fusion mostly works on elementwise operations with a linear complexity with $h$~\cite{gustafson1988reevaluating}.

Streaming, compared to the other strategies, is less effective on \texttt{Tree-LSTM} than on \texttt{Fixed-LSTM}, as we have found the depth of the input trees in SST exhibit high variance, i.e. some trees are much deeper than others. In this case,  many batching tasks only have one vertex to be evaluated. The computation is highly fragmented and the efficiency is bounded by kernel launching latency. Lazy batching and fusion still help as they both reduce kernel launches (\S\ref{sec:graph_execution}). Streaming, which tries to pipeline multiple kernels, can hardly yield obvious improvement.

%fewer nodes near the root, and 
%. In this case, the overall computation is more 
%In Figure \ref{fig:ablation}, we can see the speedup of fusion optimization becomes smaller with larger hidden size (h), while the lazy batching and Streaming optimizations are just the opposite. 
%This is because the fusion optimization is applied to element-wise operations while lazy batching and streaming mainly benefit the matrix-multiplication(MM) operation in the LSTM vertex function. The algorithm complexity of MM is O($h^2$) and that of the element-wise operations are O($h$). So the fusion operations accounts for a smaller part in the total execution time with the increase of the hidden size (h). According to the Amdahl's law\cite{gustafson1988reevaluating}, lazy batching and streamming optimization benefit more when h is large. 
%For the Tree-LSTM on SST dataset(Figure. \ref{fig:ablation} (b)), we find in most task-ids, there is only one job. It implies a great variance in the model structure between samples. In this case, the overall computation is more latency(kernel launch) bounded than computation bounded. The lazy batching and fusion optimizations benefit as well because they aggregates small operations into a larger one. However, the streamming optimization which is to overlap two kernels, can hardly contribute.

%Without discrimination, we will treat this overhead as the graph construction overhead of \texttt{Fold}. Though it is not in the same sense as the graph construction overhead in dynamic declaration-based strategies, it is a necessary step within \texttt{Fold}'s workflow and takes a significant amount of time. We'll discuss this overhead in section~\cite{sec:ablation}
\noindent \textbf{Memory Management.} 
Cavs' performance advantage also credits to its better memory management that reduces memory movements while guarantees memory continuity. 

Quantitatively, it is difficult to compare Cavs to Fold, as Fold relies on TensorFlow where memory management is highly coupled with other system aspects. Qualitatively, we find Cavs requires less memory movement (e.g. memcpy) during dynamic batching. Built upon the \texttt{tf\_while} operator, whenever Fold performs depth-based batching at $d$, it has to move all the contents of nodes in the dataflow graphs at depth $d-1$ to a desired location, as the control flow does not support cross-depth memory indexing. This results in redundant memcpy, especially when the graphs are highly skewed. By contrast, Cavs only copies contents that are necessary to the batching task.
DyNet has a specialized  memory management strategy for dynamic NNs. Compared to Cavs, it however suffers substantial overhead caused by repeated checks of the memory continuity -- whenever DyNet wants to batch operators with same signatures, it checks whether their inputs are continuous on memory~\cite{neubig2017fly}. The checking overhead increases with $bs$ and is more prominent on GPUs. Thanks to the simplicity of both systems, we are able to profile the memory-related overhead during both training and inference, and separate it from computation. We compare them on \texttt{TreeLSTM}, and report the breakdown time per epoch in Table~\ref{tab:memory_ablation} under different $bs$. We observe the improvement is significant (2x - 3x) at larger $bs$, especially during inference where DyNet has its continuity checks concentrated.

\begin{table}[!tb]
\scriptsize
\centering
%\vspace{-10pt}
\begin{tabular}{ |K{0.25cm}|K{1.1cm}|K{1.1cm}|K{0.9cm}|K{1.1cm}| } 
%\begin{tabular}{ |K{0.4cm}|c|c|c|c| } 
\hline
\multirow{2}{*}{$bs$} & \multicolumn{2}{K{2.2cm}|}{Memory operations (s) (\emph{Cavs / DyNet})} & \multicolumn{2}{K{2.2cm}|}{Computation (s) (\emph{Cavs / DyNet})} \\
\cline{2-5}
& Train & Inference &  Train & Inference \\
\hline
\hline
%0.57472, 1.05728, 2.01856, 3.88992, 7.961600000000001, 15.756
%3.0976, 3.93472, 6.15424, 10.6368, 18.4832, 32.4352
%4.10496, 7.974399999999999, 16.0, 33.664, 70.6048, 152.96
16   &  \textbf{1.14} / 1.33 & \textbf{0.6} / 1.33  &  \textbf{9.8} / 12  &  \textbf{2.9} / 8.53 \\
32   &  \textbf{0.67} / 0.87 & \textbf{0.35} / 0.87 & \textbf{6.1} / 9.8  & \textbf{1.9} / 5.35\\
64   &  \textbf{0.39} / 0.6 & \textbf{0.21} / 0.6 & \textbf{4.0} / 7.4 &  \textbf{1.3} / 3.48  \\
128  &  \textbf{0.25} / 0.44 & \textbf{0.13} / 0.44  & \textbf{2.9} / 5.9 & \textbf{0.97} / 2.52 \\
256  &  \textbf{0.17} / 0.44 &  \textbf{0.09} / 0.44 & \textbf{2.3} / 5.4 &  \textbf{0.77} / 2.58\\
\hline
\end{tabular}
\vspace{-5pt}
\caption{\small The breakdowns of the average time per epoch on memory-related operations and computation. We compare Cavs to DyNet on \texttt{Tree-LSTM} on training and inference, with varying $bs$.}
\vspace{-10pt}
\label{tab:memory_ablation}
\end{table}

\noindent \textbf{Others.}
Despite system advantages, we also try to investigate whether Cavs, as an interface, simplifies user programs (though we do not claim as a contribution). %and intellectually eases their programming of dynamic NNs.
We compare Cavs to Fold and DyNet in terms of the lines of code (LoC) needed to create a few notable dynamic NNs, including \texttt{Var-LSTM}, \texttt{Tree-LSTM}, and multi-layer sequence LSTM, with Python as the host language. If only for model declaration, Fold in general has 3.5x more LoC than Cavs, while DyNet has slightly more LoC than Cavs because of the function to repeatedly declare graphs.

\vspace{-5pt}
\section{Related Work}
\label{sec:related_work}
%\gn{The intro and explanation actually does quite a good job of covering related work. Perhaps this section could be merged into the relevant sections in the intro etc. to save space?}
%\hao{Qirong: qirong thinks we should add some discussion on graph computing literature, such as pregel, naiad, powergraph, etc. A single sentence could be enough. I think we can remove the discussion on Caffe, TF, as we have discussed them in Figure.2. We can replace them with graph computing literature.}

%First, graph computing carries on the computation on a big natural graph while dynamic neural networks faces large numbers of small graphs(one graph per sample). So batched computation and data-flow graph optimizations are the major concerns in the design of systems for dynamic neural networks, instead of the graph partition in graph computing. Second, the whole dynamic neural network may comprise several dynamic components(e.g. multi-layer RNN), each of which is a graph by itself. Two types of edges, intra-graph edges and inter-graph edges, exist in the dynamic neural networks, while there are only intra-graph edges in graph computing. We introduce the ``push'' and ``pull'' to represent the connection of different graphs.
In addition to \S\ref{sec:background}, we discuss some other related works.

\noindent \textbf{Graph execution optimization.}
Optimizing the execution of DL dataflow graphs comes in mainly two ways: better operator implementations or optimizing the execution of (sub-)graphs. As Cavs is implemented as a plug-in to enhance existing frameworks, it benefits from any improved implementations of specific operators (e.g. cuDNN)~\cite{grave2016efficient,MKL-DNN, chetlur2014cudnn, eigenweb}. In addition, Cavs has optimized implementations for its proposed four primitives (\texttt{gather}/\texttt{scatter}/\texttt{pull}/\texttt{push}).
%For the former, existing research includes the highly efficient implementation of convolution kernel~\cite{vasilache2014fast, li2016optimizing}, softmax kernel~\cite{grave2016efficient}, and DNN/blas operator toolkits~\cite{MKL-DNN, chetlur2014cudnn, eigenweb}. Our work is built on existing libraries such as cuDNN and cuBLAS while provides a manual optimized implementation for the four primitives (gather/scatter/pull/push).
At the graph level, a variety of well-developed techniques from other areas, such as kernel fusion, common subexpression elimination, and constant folding, have been adapted and applied on speeding the computation of DL dataflow graphs~\cite{abadi2016tensorflow,chen2015mxnet,caffe2, xla}. They are usually incorporated after the graph declaration, but before the execution, so that the actual computation is conducted on an optimized graph other than the original one. 
%For the latter, standard cross operator optimizations have been applied in the state of the art DL frameworks such as 
%kernel fusion, common subexpression elimination, constant folding (in Tensorflow~\citep{abadi2016tensorflow}) and streamming (also called DAG-net in caffe2's~\cite{caffe2} terminology). 
%Moreover, manual kernel fusion has been more detailed analyzed in the High Performance Computing(HPC) area~\cite{wahib2014scalable} and applied in graph computing~\cite{wang2016gunrock}. Moreover generalized forms of these optimizations are implemented by transforming the Intermediate Representation(IR), and has been realized in existing compiler frameworks such as LLVM~\cite{lattner2004llvm}, Rose Compiler Framework~\cite{quinlan2000rose}, Halide~\cite{ragan2013halide}, and TensorFlow-XLA~\cite{xla}. 
However, these graph optimizations are less beneficial in dynamic declaration, in which the graph changes with the sample, and needs to be re-processed and re-optimized every iteration, and may cause substantial overhead.
%structure may change and need to be optimized once and evaluated many times. For dynamic declaration, 
On the contrary, Cavs separates the static vertex function from the dynamic-varying input graph, so it benefits from most of the aforementioned optimizations, as we have shown in \S\ref{sec:ablation}. We draw insights from these strategies and reflect them in Cavs' execution engine. We further propose lazy batching and streaming to exploit more parallelism exposed by our programming model.
%these optimizations are orthogonal to Cavs' vertex-centric design, which separates the vertex function from the graph structure and makes the graph optimization beneficial in dynamic data-flow graph area. Specifically, our lazy batching and streaming are based on the proposed primitives because they reveals the intra- and inter- vertex dependency and potential parallelism.

\noindent \textbf{Vertex-centric models.}
The vertex-centric programming model has been extensively developed in the area of graph computing~\cite{malewicz2010pregel, gonzalez2012powergraph,chen2015powerlyra, sundaram2015graphmat}. Cavs draws insights from the GAS model~\cite{gonzalez2012powergraph}, but faces totally different challenges in system and interface design, such as expressiveness, scheduling for batched execution of different graphs, guaranteeing the memory continuity, etc., as we have discussed in \S\ref{sec:api}.

%\vspace{-10pt}
\section{Conclusion}
We present Cavs as a vertex-centric programming interface as well as  an efficient system for dynamic deep learning. Cavs represents a dynamic NN structure as static vertex functions and dynamic input graphs. It provides four novel APIs to allow users to easily program these types of NNs. With designed scheduling policy, memory management strategy, and graph execution optimizations, Cavs avoids substantial graph construction overhead suffered by dynamic declaration, and reports new state-of-the-art system performance for various notable dynamic NN architectures.

\begin{comment}
%% Acknowledgments
\begin{acks}                            %% acks environment is optional
                                        %% contents suppressed with 'anonymous'
  %% Commands \grantsponsor{<sponsorID>}{<name>}{<url>} and
  %% \grantnum[<url>]{<sponsorID>}{<number>} should be used to
  %% acknowledge financial support and will be used by metadata
  %% extraction tools.
  This material is based upon work supported by the
  \grantsponsor{GS100000001}{National Science
    Foundation}{http://dx.doi.org/10.13039/100000001} under Grant
  No.~\grantnum{GS100000001}{nnnnnnn} and Grant
  No.~\grantnum{GS100000001}{mmmmmmm}.  Any opinions, findings, and
  conclusions or recommendations expressed in this material are those
  of the author and do not necessarily reflect the views of the
  National Science Foundation.
\end{acks}
\end{comment}

%% Bibliography
\bibliographystyle{acm}
\bibliography{cavs}

%Text of appendix \ldots

\end{document}